%% file: main.tex
\documentclass{article}

\usepackage{microtype}
\usepackage{graphicx}
\usepackage{subfigure}
\usepackage{booktabs} 

\usepackage{hyperref}

\usepackage[accepted]{icml2023}

\usepackage{amsmath}
\usepackage{amssymb}
\usepackage{mathtools}
\usepackage{amsthm}
\usepackage{enumitem}
\usepackage{xspace}
\usepackage{caption}
\usepackage[capitalize,noabbrev]{cleveref}
\usepackage{bbm}

\theoremstyle{plain}
\newtheorem{theorem}{Theorem}[section]
\newtheorem{proposition}[theorem]{Proposition}

\theoremstyle{definition}

\theoremstyle{remark}

\usepackage[disable,textsize=tiny]{todonotes}

\input{mathcommand.tex}

\newcommand{\method}{\textsc{ALT-OPT}\xspace}
\icmltitlerunning{Alternately Optimized Graph Neural Networks}

\begin{document}

\twocolumn[
\icmltitle{Alternately Optimized Graph Neural Networks}




\begin{icmlauthorlist}
\icmlauthor{Haoyu Han}{u1}
\icmlauthor{Xiaorui Liu}{u2}
\icmlauthor{Haitao Mao}{u1}
\icmlauthor{MohamadAli Torkamani}{comp2}
\icmlauthor{Feng Shi}{comp1}
\icmlauthor{Victor Lee}{comp1}
\icmlauthor{Jiliang Tang}{u1}
\end{icmlauthorlist}

\icmlaffiliation{u1}{Department of Computer Science and Engineering, Michigan State University, East Lansing, US}
\icmlaffiliation{u2}{Department of Computer Science, North
Carolina State University, Raleigh, US}
\icmlaffiliation{comp1}{TigerGraph, US}
\icmlaffiliation{comp2}{Amazon, US (this work does not relate to the author's position at Amazon)}

\icmlcorrespondingauthor{Haoyu Han}{hanhaoy1@msu.edu}

\icmlkeywords{Graph Neural Networks, Semi-supervised Node Classification, Alternating Optimization}

\vskip 0.3in
]



\printAffiliationsAndNotice{}

\begin{abstract}
\input{00abstract.tex}
\end{abstract}











\section{Introduction}
\label{sec:intro}
\input{01introducetion_JT}

\section{GNNs as Bi-level Optimization}
\label{sec:bi-level}
\input{02-0Bilevel-V2}

\section{The Proposed Framework}
\label{sec:framework}

\input{02framework}

\section{Experiment}
\label{sec:exp}
\input{03experiments}

\section{Related Works}
\label{sec:related}
\input{04related}

\section{Conclusion}
\input{05conclusion.tex}

\section*{Acknowledgements}
Haoyu Han, Haitao Mao, and Jiliang Tang are supported by the National Science Foundation (NSF) under grant numbers CNS1815636, IIS1845081, IIS1928278, IIS1955285, IIS2212032, IIS2212144, IOS2107215, and IOS2035472, the Army Research Office (ARO) under grant number W911NF-21-1-0198, the Home Depot, Cisco Systems Inc, Amazon Faculty Award, and SNAP.

\bibliography{reference}
\bibliographystyle{icml2023}

\newpage
\appendix
\onecolumn

\input{06appendix.tex}


\end{document}

%% file: mathcommand.tex
\usepackage{mathtools}

\usepackage{amsthm}
\usepackage{color}

\newcommand{\cut}[1]{{}}

\newcommand{\tA}{{\tilde{\vA}}}

\newcommand{\tL}{{\tilde{\vL}}}

\newcommand{\MLP}{{\text{MLP}}}

\newcommand{\vf}{{\mathbf{f}}}

\newcommand{\vy}{{\mathbf{y}}}

\newcommand{\vA}{{\mathbf{A}}}

\newcommand{\vD}{{\mathbf{D}}}

\newcommand{\vF}{{\mathbf{F}}}

\newcommand{\vI}{{\mathbf{I}}}

\newcommand{\vL}{{\mathbf{L}}}

\newcommand{\vP}{{\mathbf{P}}}

\newcommand{\vX}{{\mathbf{X}}}
\newcommand{\vY}{{\mathbf{Y}}}

\newcommand{\cC}{{\mathcal{C}}}
\newcommand{\cD}{{\mathcal{D}}}

\newcommand{\cL}{{\mathcal{L}}}

\newcommand{\cV}{{\mathcal{V}}}

\newcommand{\RR}{\mathbb{R}}

\newcommand{\vzero}{\mathbf{0}}

\newcommand{\tr}{{\mathrm{tr}}} 

\makeatletter
\let\@@span\span
\def\sp@n{\@@span\omit\advance\@multicnt\m@ne}
\makeatother

\DeclareMathOperator*{\argmin}{arg\,min}

\newcommand{\bc}{\begin{center}}
\newcommand{\ec}{\end{center}}

\newcommand{\bdm}{\begin{displaymath}}
\newcommand{\edm}{\end{displaymath}}

\newcommand{\beq}{\begin{equation}}
\newcommand{\eeq}{\end{equation}}

\newcommand{\bfl}{\begin{flushleft}}
\newcommand{\efl}{\end{flushleft}}

\newcommand{\bt}{\begin{tabbing}}
\newcommand{\et}{\end{tabbing}}

\newcommand{\beqn}{\begin{align}}
\newcommand{\eeqn}{\end{align}}

\newcommand{\beqs}{\begin{align*}} 
\newcommand{\eeqs}{\end{align*}}  



%% file: 00abstract.tex
Graph Neural Networks (GNNs) have greatly advanced the semi-supervised node classification task on graphs. 
The majority of existing GNNs are trained in an end-to-end manner that can be viewed as tackling a bi-level optimization problem. This process is often inefficient in computation and memory usage. In this work, we propose a new optimization framework for semi-supervised learning on graphs from a multi-view learning perspective. 
The proposed framework can be conveniently solved by the alternating optimization algorithms, resulting in significantly improved efficiency. 
Extensive experiments demonstrate that the proposed method can achieve comparable or better performance with state-of-the-art baselines while it has significantly better computation and memory efficiency. 


%% file: 01introducetion_JT.tex
Graph is a fundamental data structure that denotes pairwise relationships between entities in a wide variety of domains~\citep{wu2019comprehensive-survey,ma2021deep}. Semi-supervised node classification is one of the most crucial tasks on graphs. Given graph structure, node features, and a part of labels, the task aims to predict labels of the unlabeled nodes. In recent years, Graph Neural Networks~(GNNs) have been proven to be powerful in semi-supervised node classification~\citep{gilmer2017neural,kipf2016semi, velivckovic2017graph}.  A typical GNN model mainly contains two operators, i.e., \textit{feature transformation} and \textit{feature propagation}. The feature transformation operator encodes input features into a low dimensional space, which is typically a learnable function. The feature propagation operator exploits graph structure to propagate information to its neighbors, which usually follows the message passing scheme ~\cite{gilmer2017neural}. 

While the message passing scheme~\cite{gilmer2017neural} empowers GNNs with the superior capability of capturing both node features and graph structure information, they suffer from the scalability limitation when training the GNNs in an end-to-end way~\cite{hamilton2017inductive}.
The end-to-end training of GNNs are in fact inefficient in computation cost and memory usage. In the forward computation, features will be propagated through all propagation layers in every epoch. In the backward computation, the gradients need be back-propagated through all propagation layers. In addition, every propagation layer needs to maintain all the hidden states for back-propagation, leading to a high memory cost.

There are extensive efforts trying to improve the scalability and efficiency of GNNs, including pre-computing methods~\cite{wu2019simplifying, rossi2020sign}, post-computing methods\cite{huang2020combining}, sampling methods~\cite{hamilton2017inductive, chen2018fastgcn}, and distributed methods~\cite{chiang2019cluster,shao2022distributed}. In particular, pre-computing and post-computing methods are the most efficient ones in dealing with large-scale graphs. SGC~\cite{wu2019simplifying} and SIGN~\cite{rossi2020sign} first propagate features as a pre-processing and then an MLP is trained based on the propagated features. C\&S \cite{huang2020combining} first trains an encoder on node features and then applies correction and smoothing steps. These methods only need to propagate information once during the whole training process so they are highly scalable and efficient. However, the performance of these GNNs is often worse than the end-to-end GNNs that propagate features in every epoch, such as APPNP, especially when the labeling rate is small~\cite{duan2022comprehensive, palowitch2022graphworld}.

In this paper, we first demonstrate that the majority of aforementioned GNNs for node classification can be considered as solving a bi-level optimization problem. Instead of following the above perspective as previous GNNs, we propose a single-level optimization problem to couple the node features and graph structure information through a multi-view semi-supervised learning framework. 
This new optimization problem can be conveniently optimized in an alternating way, leading to the proposed algorithm \method. Extensive experiments demonstrate that \method not only remarkably alleviates the computational and memory inefficiency issues of end-to-end trainable GNNs but also achieves comparable or even better performance than the state-of-the-art GNNs.

%% file: 02-0Bilevel-V2.tex


\textbf{Notations.} 
We use bold upper-case letters such as $\mathbf{X}$ to denote matrices. $\mathbf{X}_i$ denotes its $i$-th row and $\vX_{ij}$ indicates the $i$-th row and $j$-th column element. We use bold lower-case letters such as $\mathbf{x}$ to denote vectors. The Frobenius norm and trace of a matrix $\mathbf{X}$ are defined as $\|\mathbf{X}\|_{F}=\sqrt{\sum_{i j} \mathbf{X}_{i j}^{2}}$ and $tr(\mathbf{X}) = \sum_{i}\mathbf{X}_{ii}$.
Let $\mathcal{G}=(\mathcal{V}, \mathcal{E})$ be a graph, where $\mathcal{V}$ is the node set and $\mathcal{E}$ is the edge set. 
$\mathcal{N}_i$ denotes the neighborhood node set for node $v_i$.  
The graph can be represented by an adjacency matrix $\mathbf{A} \in \mathbb{R}^{n \times n}$, where $\mathbf{A}_{ij}>0$ indices that there exists an edge between nodes $v_i$ and $v_j$ in $\mathcal{G}$, or otherwise $\mathbf{A}_{ij}=0$. Let $\vD = diag(d_1, d_2, \dots, d_n)$ be the degree matrix, where $d_i = \sum_{j}\mathbf{A}_{ij}$ is the degree of node $v_i$. The graph Laplacian matrix is defined as $\mathbf{L} = \mathbf{D} - \mathbf{A}$.
We define the normalized adjacency matrix as $\tA = {\mathbf{D}}^{-\frac{1}{2}} {\mathbf{A}} {\mathbf{D}}^{-\frac{1}{2}}$ and the normalized Laplacian matrix as $\tL = \vI - \tA$.
Furthermore, suppose that each node is associated with a $d$-dimensional feature $\mathbf{x}$ and we use $\mathbf{X} = [ \mathbf{x}_1, \dots, \mathbf{x}_n ]^{\top} \in \mathrm{R}^{n \times d}$ to denote the feature matrix. In this work, we focus on the node classification task on graphs. Given a graph $\mathcal{G} = \{\mathbf{A}, \mathbf{X}\}$ and a partial set of labels $\mathcal{Y}_L=\{\mathbf{y}_1, \dots, \mathbf{y}_l\}$ for node set $\cV_L=\{v_1, \dots, v_l\}$, where $\mathbf{y}_i \in \mathbb{R}^c$ is a one-hot vector with $c$ classes, our goal is to predict labels of unlabeled nodes. 
Labels of graph $\mathcal{G}$ can be represented as a label matrix $\mathbf{Y} \in \mathbb{R}^{n \times c}$, where $\mathbf{Y}_i = \mathbf{y}_i$ if $v_i \in \cV_L$ and $\mathbf{Y}_i = \vzero$ if $v_i \in \cV_U$ where $\cV_U = \cV \setminus \cV_L$. 

There are some recent works ~\citep{zhu2021interpreting, ma2021unified, yang2021attributes} that have  unified GNNs into an optimization framework. For example, it has been proven that the message passing of GNNs, such as GCN, GAT, PPNP, and APPNP, can be considered as optimizing graph signal denoising problems~\cite{ma2021unified}. Considering the GNNs for node classification tasks, there are mainly two steps: (1) Forward process that fuses both features and structure information into a low-dimensional representation; and (2) Backward process that trains the whole model through gradient backpropagation. Aforementioned optimization perspectives~\citep{zhu2021interpreting, ma2021unified, yang2021attributes} only consider the forward computation of GNNs. In this work, we highlight that many existing works can be understood and unified as a bi-level optimization problem when considering both forward and backward processes:
\begin{align}
\label{eq:bi}
\begin{aligned}
&~~~~~~~~\min_{\Theta, \Phi} \|f_\Theta(\vF_L) - \vY_L\|_F^2, \\
\text{s.t.} \quad  \vF = &\argmin_{\vF\in \RR^{n\times h}} \|g_\Phi(\vX)  -\vF\|_F^2 + \lambda~\tr(\vF^\top \vL \vF)
\end{aligned}
\end{align}
where $\vF$ is the node representations, $\Theta$ and $\Phi$ are learnable parameters, and $\lambda$ is a hyperparameter controlling the smoothness of node features.
The inner problem in the constraint is the forward process that has been unified by aforementioned frameworks~\citep{zhu2021interpreting, ma2021unified, yang2021attributes}, and the outer problem in the objective is the backward process that trains the whole model. 


For the bi-level optimization problem Eq.~\eqref{eq:bi}, GNNs usually solve the inner problem first, and then solve the outer problem.
Some GNNs, such as GCN, GAT, and APPNP, solve the inner problem every epoch with $f_\Theta(\vF)=\vF$, which are inefficient. 
We refer to GNNs that need to propagate features in every epoch as \textbf{Persistent GNNs}. Others, such as SGC, and SIGN, actually only solve the inner problem once with $g_\Phi(\vX)=\vX$, without any learnable parameters.
Thus, it is hard for them to get optimal results, which could partially help us understand their sacrificed performance in practice. We refer to GNNs that only need to propagate once as \textbf{One-time GNNs}.


%% file: 02framework.tex
In this section, we introduce a new single-level multi-view optimization problem instead of the bi-level optimization that paves us a way to advance the efficiency of GNNs.

\subsection{A General Framework}
In this work, we consider the node features, graph structure, and node labels in the node classification task as three different views of graphs. 
We propose a general single-level optimization framework for multi-view learning by using a latent variable $\vF$ to capture these three types of information:
\begin{align} \label{eq:1}
\underset{\mathbf{F}, \Theta}{\min } \
   \lambda_1\cD_X(\vX, \vF) + \cD_A(\vA, \vF) + \lambda_2\cD_Y(\vY_L,\vF_L),
\end{align}
where $\Theta$ contains all learnable parameters, $\vF$ is the introduced latent variable shared by three views, $\cD_X(\cdot, \cdot)$, $\cD_A(\cdot, \cdot)$ and $\cD_Y(\cdot, \cdot)$ are functions to model node features, graph structure, and labels, respectively.  Hyper-parameters $\lambda_1$ and $\lambda_2$ are introduced to balance these three terms. Base on this framework, we can have numerous designs:
\begin{itemize}[leftmargin=0.2in]
    \item $\cD_X$ is to map node features $\vX$ to $\vF$. In reality, we can first transform 
    $\vX$ before mapping. Thus, feature transformation methods can be applied including traditional methods such as PCA~\citep{Collins2001AGO, Shen2009PrincipalCA} and SVD~\citep{Godunov2021SingularVD}, and deep methods such as, $\MLP$ and self-attention~\citep{Vaswani2017AttentionIA}. We also have various choices of the mapping functions such as Multi-Dimensional Scaling~(MDS)~\citep{Hout2013MultidimensionalS} which preserves the pairwise distance between $\vX$ and $\vF$ and any distance measurements. 
    \item $\cD_A$ aims to impose constraints on the latent variable $\vF$ with the graph structure. Traditional graph regularization techniques can be employed. For instance, the Laplacian regularization~\citep{Yin2016LaplacianRL} is to guide a node $i$'s feature $\vF_i$ to be similar to its neighbors;  Locally Linear Embedding (LLE)~\citep{Roweis2000NonlinearDR} is to force the $\vF_i$ be reconstructed from its neighbors. Different graph signal filters~\cite{shuman2013emerging} can also be utilized.
    Moreover, modern deep graph learning methods can be applied, such as graph embedding methods~\citep{Perozzi2014DeepWalkOL, Grover2016node2vecSF} and Graph Contrastive Learning~\citep{Zhu2020DeepGC, hu2021graph}, which implicitly encodes node similarity.
    \item $\cD_Y$ establishes the connection between the latent variable $\vF_L$ and the ground truth node label $\vY_L$ for labeled nodes. It can be any classification loss function, such as the Mean Square Error and Cross Entropy Loss.  
\end{itemize}

In this work, we set the dimensions of the latent variable $\vF$ as $\mathbb{R}^{n \times c}$, which can be considered as a soft pseudo-label matrix. As an example for this framework, the following designs are chosen for these functions: (i) for $\cD_X$, we use an $\MLP$ with parameter $\Theta$ to encode the features of node $i$ as $\MLP(\vX_i; \Theta)$, and then adopt the Euclidean norm to measure the distance between $\vF_i$ and $\vX_i$ as $\| \MLP(\vX_i;\Theta)-\vF_i \|_2^2$; (ii) for $\cD_A$, Laplacian smoothness is
imposed to constrain the distance between neighboring nodes' pseudo labels $\sum_{\left(v_{i}, v_{j}\right) \in \mathcal{E}}\|\vF_i/\sqrt{d_i}-\vF_j/\sqrt{d_j}\|_{2}^{2}$; and (iii) for $\cD_Y$, we adopt Mean Square Loss $\|\vF_i-\vY_i\|_2^2$ to constraint the distance between pseudo label of a labeled node to the ground truth label.

This design leads to our \method method, and its optimization objective $\mathcal{L}$ can be written in the matrix form as:
{\small
\begin{align} \label{eq:2}
\cL=
\lambda_1\underbrace{\|\MLP(\vX)-\vF\|_F^2}_{\cD_X} + \underbrace{\tr(\vF^\top \tL \vF)}_{\cD_A} + \lambda_2 \underbrace{\|\vF_L-\vY_L\|_F^2}_{\cD_Y}.
\end{align}
}

\noindent Compared with the bi-level optimization problem in Eq.~\eqref{eq:bi}, Eq.~\eqref{eq:2} introduces a single-level optimization problem. In the following sections, we will demonstrate that optimizing Eq.~\eqref{eq:2} alternatingly leads to substantially better efficiency. 

\subsection{An Alternating Optimization Method}
\label{sec:altopt}

Due to the coupling between the latent variable $\vF$ and model parameters $\Theta$ in Eq.~\eqref{eq:2}, it can be difficult to optimize both $\vF$ and $\Theta$ simultaneously. The alternating optimization~\citep{bezdek2002some} based iterative algorithm can be a natural solution for this challenge.
Specifically, for each iteration, we first fix the model parameters $\Theta$ and update the shared latent variable $\vF$ on all three views. Then, we fix $\vF$ and update the parameters $\Theta$, which is effective in exploring the complementary characteristics of the three views. These two steps alternate until convergence. Next, we show the alternating optimization algorithm in detail.

\noindent \textbf{Update F.} Fixing MLP, we minimize $\cL$ with respect to the latent variable $\vF$ using the gradient descent method with step sizes $\eta_L$ and $\eta_U$ for labeled and unlabeled nodes, respecitvely. 
%
%
%
{
\small
\begin{align*}
&\begin{aligned}
\vF^{k+1}_L 
&= 2\eta_L \Big ((\tA\vF^k)_L + \lambda_1 \MLP(\vX_L) + \lambda_2 \vY_L \Big) \\
&~~~~~ + \Big(1-2\eta_L (\lambda_1+\lambda_2+1)\Big)\vF^k_L \\
\end{aligned} \\
&\begin{aligned}
\vF^{k+1}_U 
&= 2\eta_U \Big ((\tA\vF^k)_U + \lambda_1 \MLP(\vX_U) \Big) \\
&~~~~~ + \Big(1-2\eta_U (\lambda_1+1)\Big)\vF^k_U.
\end{aligned}
\end{align*}
}

\noindent According to the smoothness and strong convexity of the problem with respect to $\vF$, we set $\eta_L = \eta_U=\frac{1}{2(\lambda_1+\lambda_2+1)}$ to ensure the decrease of loss value $\cL$~\citep{nesterov2018lectures}, and the update becomes:
{\small
\begin{align} 
&\begin{aligned}
\label{eq:F_L}
\vF^{k+1}_L &= \frac{1}{\lambda_1+\lambda_2+1} (\tA\vF^k)_L + \frac{\lambda_1}{\lambda_1+\lambda_2+1} \MLP(\vX_L)  \\
&\qquad + \frac{\lambda_2}{\lambda_1+\lambda_2+1} \vY_L, 
\end{aligned}\\
&\begin{aligned}
\label{eq:F_U}
\vF^{k+1}_U  &= \frac{1}{\lambda_1+\lambda_2+1} (\tA\vF^k)_U + \frac{\lambda_1}{\lambda_1+\lambda_2+1} \MLP(\vX_U)  \\
& \qquad + \frac{\lambda_2}{\lambda_1+\lambda_2+1} \vF^k_U.
\end{aligned}
\end{align}
}

\noindent \textbf{Update $\Theta$.} Fixing $\vF^{k+1}$, we minimize the loss function $\cL$ with respect to MLP parameters $\Theta$:
\begin{align} 
\label{eq:mlp}
\argmin_{\Theta}\|\MLP(\vX; \Theta)-\vF^{k+1}\|_F^2,
\end{align}
which is equivalent to training the $\MLP$ with soft pseudo labels $\vF^{k+1}$ via the mean square loss. We also explore the cross-entropy loss, and Details are in Appendix~\ref{app:crossentropy}.
In summary, \method first updates $\vF$ according to Eq.~\eqref{eq:F_L} and~\eqref{eq:F_U}, and then trains $\MLP$ with updated $\vF$ being the labels. The alternating update will be repeated until convergence.

\subsection{Efficiency of \method} 
We derive an alternating optimization algorithm from the single-level optimization perspective,
which provides significantly improved computation and memory efficiency compared to existing end-to-end GNNs as discussed below. 

\noindent \textbf{Efficiency on updating $\vF$.} Updating $\vF$ is the propagation process, which can be time-consuming. Existing GNNs usually perform both propagation and model parameters update during each iteration. For \method, variable $\vF$ and model parameters $\Theta$ are optimized separately, which is more flexible. We can update $\vF$ once while training MLP multiple times, and then alternate these two steps until convergence. During the whole training process, variable $\vF$ only needs to be updated a few times, which is highly efficient.

\noindent \textbf{Efficiency on updating $\Theta$.}  For end-to-end GNNs training, both forward and backward process need to pass through all propagation layers. For \method, it only needs to train an $\MLP$ using the generated pseudo labels $\vF$. There is no gradient backpropagation through the feature propagation process (update $\vF$), so these propagation layers do not need to store the activation and gradient values, which saves a significant amount of memory and computation.
%
Moreover, our \method is well suited for the training on large-scale graphs.
because it is easy to apply stochastic optimization to train MLP using the pseudo label due to our flexible training strategy. 
This can further improve the memory and computation efficiency as proved theoretically and empirically in extensive literature of stochastic optimization~\citep{lan2020first}.

\begin{figure*}[!htp]
  \centering
  \includegraphics[width=0.85\linewidth]{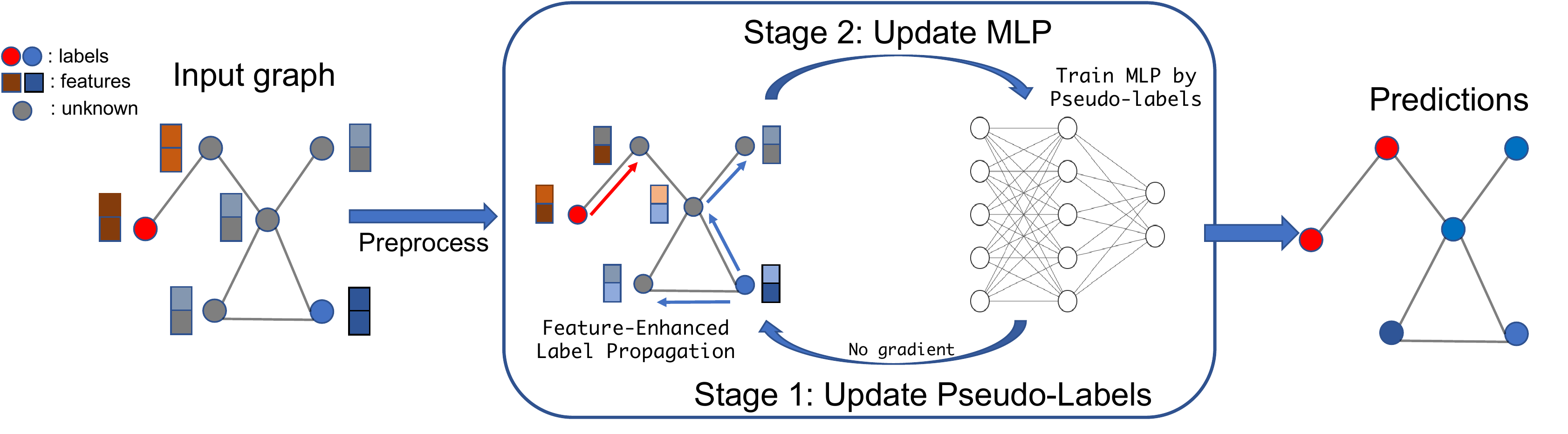}
  \caption{An overview of the proposed \method method. \method will alternately (1) generate pseudo labels without gradient; (2) train the $\MLP$ on pseudo labels. 
} 
  \label{fig:framework}
\end{figure*}

\subsection{Understandings of \method}
\label{sec:understanding}

Another important advantage of alternating optimization is that it provides helpful insights to understand \method based on the updating rules of $\vF$ and $\Theta$.

\textbf{Understanding 1: Updating $\vF$ is a feature-enhanced label propagation.} Label Propagation (LP) \citep{zhou2003learning} is a well-known graph semi-supervised learning method, which has shown great efficiency and even can work well under the low labeling rate \cite{wang2006label, karasuyama2013manifold}. 
However, LP cannot directly leverage feature information, resulting in unsatisfied performance when features are essential for downstream tasks. We provide the following proposition to compare LP and the proposed \method. 
\begin{proposition}
\label{pro:lp}
The label propagation can be written as $\text{LP}(\vY, \alpha)=\bar{\vA}\vY $, and the propagation in \method can be represented as $\vF=\bar{\vA}\left((\alpha-\beta) \MLP(\vX) + \beta \vY\right)$, where $\alpha, \beta$ are hyperparamters, $K$ is the number of propagation layers, and $\bar{\vA}=(1-\alpha)^K \tA^K+\alpha \sum_{k=0}^{K-1}(1-\alpha)^k \tA^k. $
\end{proposition}
Details can be found in Appendix~\ref{app:lp}. Proposition~\ref{pro:lp} suggests that \method propagates not only the ground truth labels, but also ``feature labels" $\MLP(\vX)$ generated by features. Thus, updating $\vF$ takes advantage of information from all three views including node features, graph structure, and labels while LP only leverages graph structure and ground-truth label information. 


\textbf{Understanding 2: Updating $\Theta$ is a pseudo-labeling approach.} Pseudo-labeling \citep{lee2013pseudo, arazo2020pseudo} is a popular method in semi-supervised learning that uses a small set of labeled data along with a large amount of unlabeled data to improve model performance. It usually generates pseudo labels for the unlabeled data and trains the deep models using both the ground truth and pseudo labels. From this perspective, \method{} uses the pseudo labels $\vF$ to train $\MLP$ such that it can achieve better performance and efficiency. We also provide the following proposition:
\begin{proposition}
\label{pro:pseudo}
If we choose to train the MLP in the proposed \method using the cross-entropy loss, the loss can be rewritten as $(\alpha - \beta) \sum_{i, j \in \cV} \bar{\vA}_{i j} CE(\vf_i, \vf_j^\prime) + \beta \sum_{i\in \cV, j \in \cV_L} \bar{\vA}_{i j} CE(\vf_i, \vy_j)$, where $\vf_i$ and $\vf_i^\prime$ are the MLP predictions of node $i$ from the current and previous iterations, respectively.
\vspace{-0.1in}
\end{proposition}
More details are in Appendix~\ref{app:pseudo}. Proposition \ref{pro:pseudo} suggests that \method uses both ground truth label $\vy_i$ and pseudo label $\vf_j^\prime$ to train the $\MLP$. A recent paper \cite{dong2021equivalence} points out that the loss for training  decouple GNNs, such as APPNP, can be represented as $\sum_{i \in \mathcal{V}, j \in \mathcal{V}_L} \frac{\bar{a}_{j i} \vf_{i, h(j)}}{\sum_{q \in \mathcal{V}} \bar{a}_{j q} \vf_{q, h(j)}}  \operatorname{CE}\left({\vf}_i, {\vy}_j\right)$, where $h(j)$ is the label of node $j$, and $\vf_{i, h(j)}$ is the $h(j)$-th entry of $\vf_i$. Compared with this loss, \method utilizes history predictions as pseudo labels, and is easier to compute as it need not to compute the denominator.




\subsection{Implementation Details of \method}

In this subsection, we detail the implementation of \method. As shown in Figure~\ref{fig:framework}, we first preprocess the node feature through a diffusion, then alternatively update pseudo label $\vF$ and $\MLP$ while taking into account the weight of pseudo labels and the class balancing problem, and finally get the predictions. Next, we describe each step in detail. 

\noindent \textbf{Preprocessing.} From Understanding 1 and 2, we use the pseudo labels generated by $\MLP$ to enhance the label propagation, so a good initialized $\MLP$ is needed. In real graphs, labeled data are usually scarce so it is challenging to get a good initialization of $\MLP$ with a small number of labels. Therefore, we first diffuse the original node features with its neighbors to get smoothing and enhanced features. The new features are obtained from $\vX^{\prime} = \text{LP}(\vX, \alpha)$. Then, we train MLP only using the labeled data for a few epochs to get an initialization, similar to pseudo-labeling methods~\citep{iscen2019label, lee2013pseudo}.

\noindent \textbf{Update F.} We initial $\vF^0=\vY$. Then we update $\vF$ for labeled nodes and unlabeled nodes by Eq.~\eqref{eq:F_L} and Eq.~\eqref{eq:F_U}, respectively. Since $\vF$ acts as pseudo labels when training the $\MLP$, we normalize $\vF$ to be the distribution of classes by the softmax function with temperature after the update: $\vF_{ij} = \frac{\exp \left(\vF_{ij} / \tau\right)}{\sum_{k=1}^{c} \exp \left(\vF_{ik} / \tau\right)}$,
where $\tau$ is a hyperparameter to control the smoothness of pseudo labels. 

\noindent \textbf{Pseudo-labels Generation.} Directly using all pseudo labels to train $\MLP$ is not appropriate due to the following reasons. First, not all pseudo labels have the same certainty. Second, pseudo-labels may not be balanced over classes, which will impede learning. 
To address the first issue, we assign a confidence weight to each pseudo-label~\citep{rizve2021defense, iscen2019label}. According to information theory, entropy can be used to quantify a distribution's uncertainty, so we define the weight for unlabeled nodes as $w_i = 1 - \frac{H(\vF_i)}{log(c)}$,
where $w_i \in [0, 1]$ and $H(\vF_i) = -\sum_{j=1}^{c} \vF_{ij} \log \vF_{ij}$ is the entropy of the pseudo label $\vF_i$. To deal with the class imbalance problem, we train the MLP using the same number of unlabeled nodes for each class with the highest weights. 

\noindent \textbf{Update MLP.} We train $\MLP$ using both labeled nodes $\cV_L$ and high confidence unlabeled nodes $\cV_{U_t}$ with the loss: 
{\small
\begin{align*}
    \sum_{i\in \cV_L} \ell(\MLP(\vX'_i; \Theta), \vF_i) + \sum_{j\in \cV_{Ut}} w_j \cdot \ell(\MLP(\vX'_j; \Theta), \vF_j),
\end{align*}
}
where $\ell(\MLP(\vX'_i; \Theta), \vF_i)=\| \MLP(\vX'_i; \Theta) - \vF_i\|_2^2$ is a MSE loss and $\Theta$ is the parameters of $\MLP$.
 
\noindent \textbf{Prediction.} The inference of the proposed method is based on the pseudo labels $\vF$, and the predicted class for the unlabeled node $i$ can be obtained as $c_i = \arg \max _{j} \vF_{ij}$.

It is important to note that the normalization of $\vF$ and pseudo-label reweighting can be incorporated into the unified framework. Please refer to Appendix~\ref{app:incoporate} for more details. The overall algorithm and implementation details~\footnote{https://github.com/haoyuhan1/ALT-OPT/} of \method are shown in Appendix~\ref{app:algorithm}.

\subsection{Complexity Analysis}
\label{sec:complexity}
We provide time and memory complexity analyses for \method and the following representative GNNs: GCN~\citep{kipf2016semi}, SGC~\citep{wu2019simplifying}, and APPNP~\citep{klicpera2018predict}.

Suppose that $p$ is the number of propagation layers, $n$ is the number of nodes, $m$ is the number of edges, and $c$ is the number of classes.  For simplicity, we assume that the hidden feature dimension is a fixed $d$ for all transformation layers, and we have $c \ll d$ in most cases;  all feature transformations are updated $t$ epochs. The adjacent matrix $\vA$ is a sparse matrix, and forward and backward propagation have the same cost.  Following \citep{li2021training}, we only analyze the inherent differences across models by assuming that they have the same transformation layers ($\MLP$), allowing us to disregard the time and memory footprint of $\MLP$.
The time and memory complexities are summarized in Table~\ref{tab:com}.

\textbf{Time complexity.} We first analyze the time complexity of feature aggregation. The feature aggregation can be implemented as a sparse-dense matrix multiplication with cost $O(md)$ if the feature has $d$ dimensions. Therefore, the time complexity of training a $p$-layer GCN for $t$ epochs is $O(2tpmd)$ with the gradient backpropagation. For SGC, we only need $p$ steps of feature propagation, so the time complexity is $O(pmd)$. For APPNP, the gradient also needs to backpropagate through $p$ layers, but the feature dimension is $c$,  resulting in the time complexity of $O(2tpmc)$. Regarding \method, as the model are optimized in an alternating way, there is no need to do both feature transformation and aggregation in each epoch. Rather, we can propagate the pseudo labels only for $k$ times during the whole training process. As a result, the time complexity of \method is $O(kpmc)$. In practice, choosing $k$ from $\{2,3,4, 5\}$ can achieve very promising performance, while $t$ needs to be 500 or 1,000 for other models to converge.
\begin{table}[htb]
\centering
\caption{Comparison of time and memory complexities.}
\label{tab:com}
\begin{tabular}{c|cc}
\hline
Method       & \multicolumn{1}{c}{Time} & \multicolumn{1}{c}{Memory} \\ \hline
GCN          & $O(2tpmd)$    & $O(nd+pnd)$                   \\
SGC          & $O(pmd)$      & $O(nd)$                    \\
APPNP        & $O(2tpmc)$    & $O(nd+pnc)$                   \\
\method      & $O(kpmc)$     & $O(nd+nc)$                   \\ \hline
\end{tabular}
\vspace{-0.1in}
\end{table}

\textbf{Memory complexity.} All models require $O(nd)$ memory for storing node features. For the end-to-end training models, we need to store the intermediate state at each layer for gradient calculation. Specifically, for GCN, we need to store the hidden state for $p$ layers, so the memory complexity is $O((p+1)nd)$. SGC only needs to store the propagated feature $O(nd)$ as we omit the memory of network parameters.
Similarly, APPNP has the memory complexity of $O(nd+pnc)$.
As for \method, it does not need to store the gradients at each propagation layer. Instead, \method needs to hold the pseudo label $\vF$. So the memory complexity of \method is $O(nd+nc)$. 

If we omit the difference in the dimension of the propagation features (d = c), the time and memory of GCN and APPNP are the same, as they are both Pesistent GNNs that require feature propagation in each epoch. Similarly, the One-time GNNs that only need to propagate once, such as SGC, SIGN, and C\&S have the same time and memory complexity. \method is a \textit{Lazy 
Propagation method} since the features are propagated $k$ times during training with $k$ being a small number. Thus, \method can be seen as a balance between these two groups of methods.

%% file: 03experiments.tex
In this section, we verify the effectiveness of the proposed \method by comprehensive experiments. In particular, we try to answer the following questions:
\vspace{-0.1in}
\begin{itemize}[leftmargin=0.2in]
\setlength\itemsep{-0.2em}
    \item \textbf{RQ1:} How does \method perform when compared to other baseline models? 
    \item \textbf{RQ2:} Is \method more time and memory efficient than state-of-the-art GNNs?
    \item \textbf{RQ3:} How do different components affect \method?
\end{itemize}

\subsection{Experimental settings}
\textbf{Datasets.} 
For the transductive semi-supervised node classification task, we choose nine commonly used datasets including three citation datasets, i.e., Cora, Citeseer and Pubmed \citep{sen2008collective}, two coauthors datasets, i.e., CS and Physics, two Amazon datasets, i.e., Computers and Photo \citep{shchur2018pitfalls}, and two OGB datasets, i.e., ogbn-arxiv and ogbn-products \citep{hu2020open}. For the inductive node classification task, we use Reddit and Flikcr datasets \citep{zeng2019graphsaint}. We also test the proposed method on two heterophily graphs, i.e., Chameleon and Squirrel~\cite{pei2020geom}. More details about these datasets are shown in Appendix~\ref{app:dataset}.

We use 10 random data splits for all the datasets except the OGB datasets. For the OGB datasets, we use the fixed data split. We run the experiments 3 times for each split and report the average performance and standard deviation. Besides, we also test multiple labeling rates, i.e., low label rates with 5, 10, 20, and 60 labeled nodes per class and high label rates with 30\%  and 60 \% labeled nodes per class, to get a comprehensive comparison.

\textbf{Baselines.} We compare the proposed \method with three groups of methods: (i) Persistent GNNs, i.e.,  GCN~\citep{kipf2016semi}, GAT~\citep{velivckovic2017graph} and APPNP~\citep{klicpera2018predict}; (ii) One-time GNNs, i.e., SGC~\citep{wu2019simplifying}, SIGN~\cite{rossi2020sign}, and C\&S \citep{huang2020combining}; and (iii) Non-GNN methods including MLP and Label Propagation~\citep{zhou2003learning}. We report the test accuracy selected by the the best validation accuracy.
Parameter settings are summarized in Appendix~\ref{app:parameter}.

\subsection{Performance Comparison on Benchmark Datasets}
\subsubsection{Transductive Node Classification.} 

The transductive node classification results are partially shown in Table \ref{tab:result1}. We leave results on more datasets and methods with more learning rate setting in Appendix~\ref{app:transductive} due to the space limitation. From these results, we can make the following observations:
\vspace{-0.1in}
\begin{itemize}[leftmargin=0.2in]
\setlength\itemsep{-0.2em}
    \item \method consistently outperforms other models at low label rates on all datasets. For example, in Cora and CiteSeer with label rate 5, our method can gain $1.2\%$ and $5.6\%$ relative improvement compared to the best baselines.
    This is because the pseudo labels generated by our framework are helpful for training models when there are few labels available.
    When the label rate is high, our method is also comparable to the best results. In addition, \method is alternately optimized, which suggests that end-to-end training could not be necessary for node semi-supervised classification.
    \item  \method performs the best on two OGB datasets. For example, in the large ogbn-products dataset, it obtains 7.86\% and 2.63\% relative improvement compared to APPNP and SIGN, respectively.
    \item Compared with the One-time GNNs, Persistent GNNs usually perform better when the labeling rate is low. In addition, the label propagation outperforms $\MLP$ in most cases, indicating the rationality of our proposed feature-enhanced label propagation.
    \item The standard deviation of all models is not small across different data splits, especially when the label rate is very low. It demonstrates that splits can significantly affect a model's performance. A similar finding is also observed in the PyTorch-Geometric paper~\citep{fey2019fast}.
    \vspace{-0.2in}
\end{itemize}

\begin{table*}[!htb]
\caption{Transductive node classification accuracy (\%) on benchmark datasets.} 
\label{tab:result1}
\resizebox{\textwidth}{!}{%
\begin{tabular}{cc|ccc|ccc|c}
\hline
\multicolumn{2}{c|}{Method}                     & \multicolumn{3}{c|}{Persistent GNNs}                  & \multicolumn{3}{c|}{One-time GNNs}                         & Ours                      \\ \hline
\multicolumn{1}{c|}{Dataset}       & Label & GCN              & GAT              & APPNP                     & SGC              & SIGN         & C\&S                      & \method                   \\ \hline
\multicolumn{1}{c|}{Cora}          & 5          & 70.68 $\pm$ 2.17 & 72.97 $\pm$ 2.23 & 75.86 $\pm$ 2.34          & 70.06 $\pm$ 1.95 & 69.81 $\pm$ 3.13 & 56.52 $\pm$ 5.53          & \textbf{76.78 $\pm$ 2.56} \\
\multicolumn{1}{c|}{}              & 10         & 76.50 $\pm$ 1.42 & 78.03 $\pm$ 1.17 & 80.29 $\pm$ 1.00          & 76.28 $\pm$ 1.22 & 76.25 $\pm$ 1.26 & 71.04 $\pm$ 3.30          & \textbf{80.66 $\pm$ 1.92} \\
\multicolumn{1}{c|}{}              & 20         & 79.41 $\pm$ 1.30 & 81.39 $\pm$ 1.41 & 82.34 $\pm$ 0.67          & 80.30 $\pm$ 1.72 & 79.71 $\pm$ 1.11 & 77.96 $\pm$ 2.13          & \textbf{82.66 $\pm$ 0.98} \\
\multicolumn{1}{c|}{}              & 60         & 84.30 $\pm$ 1.44 & 85.11 $\pm$ 1.10 & 85.49 $\pm$ 1.25          & 84.17 $\pm$ 1.39 & 84.16 $\pm$ 1.18 & 82.21 $\pm$ 1.45          & \textbf{85.60 $\pm$ 1.12} \\
\multicolumn{1}{c|}{}              & 30\%       & 86.87 $\pm$ 1.35 & 87.24 $\pm$ 1.19 & \textbf{87.77 $\pm$ 1.13} & 86.97 $\pm$ 0.90 & 87.17 $\pm$ 1.28 & 87.60 $\pm$ 1.12          & 87.70 $\pm$ 1.19          \\ \hline
\multicolumn{1}{c|}{CiteSeer}      & 5          & 61.27 $\pm$ 3.85 & 62.60 $\pm$ 3.34 & 63.92 $\pm$ 3.39          & 60.21 $\pm$ 3.48 & 57.44 $\pm$ 3.71 & 50.39 $\pm$ 4.70          & \textbf{67.48 $\pm$ 2.90} \\
\multicolumn{1}{c|}{}              & 10         & 66.28 $\pm$ 2.14 & 66.81 $\pm$ 2.10 & 67.57 $\pm$ 2.05          & 65.23 $\pm$ 2.36 & 63.87 $\pm$ 3.09 & 58.96 $\pm$ 2.75          & \textbf{69.39 $\pm$ 2.59} \\
\multicolumn{1}{c|}{}              & 20         & 69.60 $\pm$ 1.67 & 69.66 $\pm$ 1.47 & 70.85 $\pm$ 1.45          & 68.82 $\pm$ 2.11 & 68.60 $\pm$ 1.94 & 65.85 $\pm$ 2.74          & \textbf{71.26 $\pm$ 1.69} \\
\multicolumn{1}{c|}{}              & 60         & 72.52 $\pm$ 1.74 & 73.10 $\pm$ 1.20 & \textbf{73.50 $\pm$ 1.54} & 71.43 $\pm$ 1.26 & 72.63 $\pm$ 1.39 & 71.21 $\pm$ 1.79          & 72.84 $\pm$ 1.65          \\
\multicolumn{1}{c|}{}              & 30\%       & 75.20 $\pm$ 0.85 & 75.01 $\pm$ 0.99 & \textbf{75.71 $\pm$ 0.71} & 75.09 $\pm$ 1.01 & 74.44 $\pm$ 0.83 & 74.65 $\pm$ 0.95          & 75.09 $\pm$ 0.79          \\ \hline
\multicolumn{1}{c|}{Pubmed}        & 5          & 69.76 $\pm$ 6.46 & 70.42 $\pm$ 5.36 & 72.68 $\pm$ 5.68          & 68.55 $\pm$ 6.88 & 66.52 $\pm$ 6.15 & 65.3 $\pm$ 6.02           & \textbf{73.51 $\pm$ 4.80} \\
\multicolumn{1}{c|}{}              & 10         & 72.79 $\pm$ 3.58 & 73.35 $\pm$ 3.83 & 75.53 $\pm$ 3.85          & 72.80 $\pm$ 3.55 & 71.32 $\pm$ 3.70 & 72.51 $\pm$ 3.75          & \textbf{75.55 $\pm$ 5.09} \\
\multicolumn{1}{c|}{}              & 20         & 77.43 $\pm$ 1.93 & 77.43 $\pm$ 2.66 & 78.93 $\pm$ 2.11          & 76.48 $\pm$ 2.84 & 76.39 $\pm$ 2.65 & 75.34 $\pm$ 2.49          & \textbf{79.16 $\pm$ 2.26} \\
\multicolumn{1}{c|}{}              & 60         & 82.00 $\pm$ 1.62 & 81.40 $\pm$ 1.40 & \textbf{82.55 $\pm$ 1.47} & 80.34 $\pm$ 1.61 & 81.75 $\pm$ 1.55 & 80.63 $\pm$ 1.49          & 82.53 $\pm$ 1.76          \\
\multicolumn{1}{c|}{}              & 30\%       & 88.07 $\pm$ 0.29 & 86.51 $\pm$ 0.41 & 87.56 $\pm$ 0.39          & 86.23 $\pm$ 0.43 & \textbf{89.09 $\pm$ 0.33} & 88.44 $\pm$ 0.40 & 88.24 $\pm$ 0.36          \\ \hline
\multicolumn{1}{c|}{CS}            &    20        & 91.73 $\pm$ 0.49     & 90.96 $\pm$ 0.46     & 92.38 $\pm$ 0.38              & 90.32 $\pm$ 0.99     & 92.02 $\pm$ 0.41 & 92.41 $\pm$ 0.44     & \textbf{92.77 $\pm$ 0.50 }              \\ \hline
\multicolumn{1}{c|}{Physics}       &      20    &  93.29 $\pm$ 0.80     & 92.81 $\pm$ 1.03     & 93.49 $\pm$ 0.67              & 93.23 $\pm$ 0.59     & 93.03 $\pm$ 1.15 & 93.23 $\pm$ 0.55     & \textbf{94.63 $\pm$ 0.31}              \\ \hline
\multicolumn{1}{c|}{Computers}     &     20      & 79.17 $\pm$ 1.92     & 78.38 $\pm$ 2.27     & 79.07 $\pm$ 2.34              & 73.00 $\pm$ 2.0      & 73.04 $\pm$1.15  & 73.25$\pm$ 2.09      & \textbf{79.12 $\pm$ 2.50}              \\ \hline
\multicolumn{1}{c|}{Photo}         &     20      & 89.94 $\pm$ 1.22     & 89.24 $\pm$ 1.42     & 90.87 $\pm$ 1.14              & 83.50 $\pm$ 2.9      & 86.11 $\pm$ 0.66 & 84.87 $\pm$ 1.04     & \textbf{91.23 $\pm$ 1.26 }             \\ \hline
\multicolumn{1}{c|}{ogbn-arxiv}    & 54\%       & 71.91 $\pm$ 0.15 & 71.92 $\pm$ 0.17 & 71.61 $\pm$ 0.30          & 68.74 $\pm$ 0.12 & 71.95 $\pm$ 0.11 & 71.03 $\pm$ 0.15          & \textbf{72.76 $\pm$ 0.17}     \\ \hline
\multicolumn{1}{c|}{ogbn-products} & 8\%        & 75.70 $\pm$ 0.19     & OOM              & 76.62 $\pm$ 0.13              &       74.29 $\pm$ 0.12  & 80.52$\pm$0.16   &      77.11 $\pm$ 0.06           & \textbf{82.64$\pm$ 0.21}      \\ \hline
\end{tabular}
}
\end{table*}

\subsubsection{Inductive Node Classification.} 
For inductive node classification, only training nodes can be observed in the graph during training, and all nodes can be used during the inference~\citep{zeng2019graphsaint}. For \method, we first train an $\MLP$ with the original features and then do inference for the unlabeled node using the feature-enhanced label propagation in Eq~\eqref{eq:F_L} and Eq~\eqref{eq:F_U}.

\begin{figure}[htb]
    \centering
    \includegraphics[width=0.8\linewidth, height=0.5\linewidth]{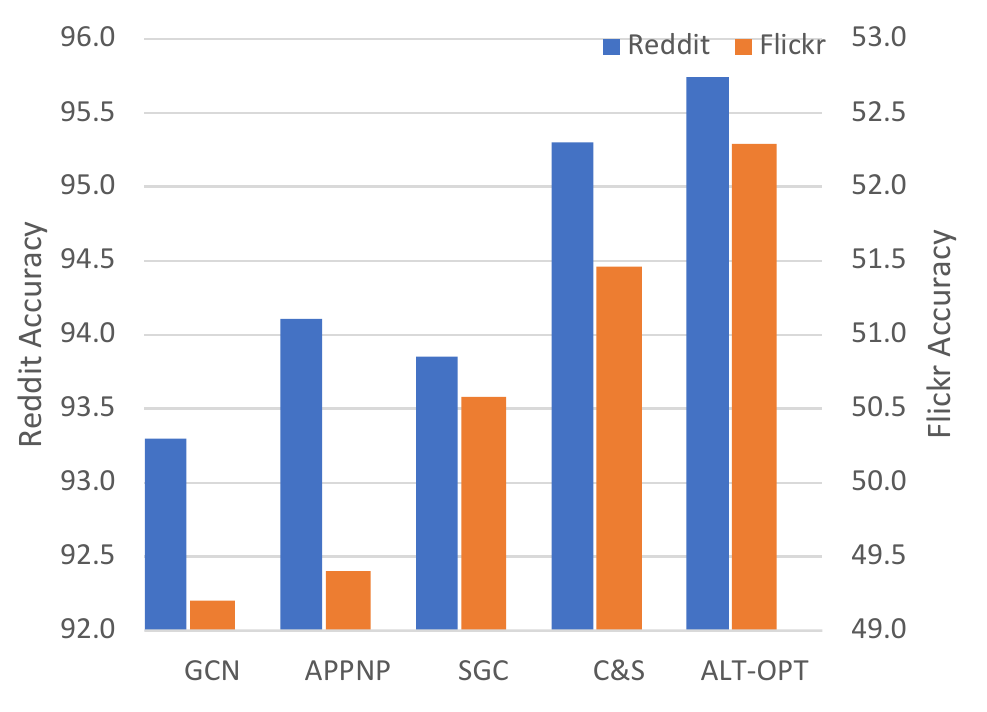}
    \vspace{-0.1in}
    \caption{Inductive node classification accuracy (\%).}
    \label{fig:indcuctive}
    \vspace{-0.1in}
\end{figure}

As shown in Figure~\ref{fig:indcuctive} and Appendix~\ref{app:inductive}, our \method outperforms other baselines on the inductive node classification task. The only difference between \method and $\MLP$ is the feature-enhanced label propagation, but our method can achieve 52.3\% and 38.1 \% relative improvement compared to  MLP. The performance improvement can demonstrate the superiority of our feature-enhanced label propagation.

\subsubsection{Performance on Heterophily Graphs.}
\method is a specific instance of the proposed single-level optimization framework. It utilizes graph Laplacian regularization, expressed as $\tr(\vF^\top\tL\vF)$, to constrain node features with the graph structure. The Laplacian regularization is a low-pass filter, potentially limiting its performance with heterophily graphs, where high-frequency signals are helpful.

Nonetheless, the proposed Eq.~\eqref{eq:1} serves as a general framework, permitting the incorporation of different filters to seize the high-frequency signals prevalent in heterophily graphs. To illustrate this, we introduce a novel regularization term, denoted as $\cD_A = \tr(\vF^\top\tL^2\vF)$, which is capable of capturing both low-pass and high-pass graph signals. Consequently, this results in a new method, which we term ALT-OPT-H for brevity. The specifics of this method are as follows:

{\small
\begin{align*} \label{eq:heter}
\cL=
\lambda_1\underbrace{\|\MLP(\vX)-\vF\|_F^2}_{\cD_X} + \underbrace{\tr(\vF^\top \tL^2 \vF)}_{\cD_A} + \lambda_2 \underbrace{\|\vF_L-\vY_L\|_F^2}_{\cD_Y},
\end{align*}
} 
which can be solved using the same alternating optimization method as in Equation~\eqref{eq:2}. We carry out experiments on two of the most widely utilized heterophily datasets, namely Chameleon and Squirrel. We maintain the same settings as in the study by \citet{lim2021large}, and compare several renowned heterophily GNNs, such as Geom-GCN~\citep{pei2020geom}, H$_2$GCN~\citep{zhu2020beyond}, MixHop~\citep{abu2019mixhop}, GCNII~\citep{chen2020simple}, GPR-GNN~\citep{chien2020adaptive}, and LINKX~\cite{lim2021large}. 
The results 
in Table~\ref{tab:hetero} show that our ALT-OPT-H method displays competitive performance on heterophily datasets, necessitating only a slight adjustment to the graph filter. This attests to the flexibility of the proposed framework (Eq.~\eqref{eq:1}), which can perform well on both homophily and heterophily graphs.

\begin{table}[!htb]
\centering
\caption{Performance on the heterophily graphs.}
\label{tab:hetero}
\begin{tabular}{c|cc}
\hline
Dataset   & Chameleon    & Squirrel     \\ \hline
Geom-GCN  & 60.90        & 38.14        \\
H$_2$GCN  & 59.39 ± 1.98 & 37.90 ± 2.02 \\
MixHop    & 60.50 ± 2.53 & 43.80 ± 1.48 \\
GCNII     & 62.48        & 56.63 ± 1.17 \\
GPR-GNN   & 62.85 ± 2.90 & 54.35 ± 0.87 \\
LINKX     & 68.42 ± 1.38 & \textbf{61.81 ± 1.80} \\
ALT-OPT-H & \textbf{70.62 ± 1.93} & 61.56 ± 1.81 \\ \hline
\end{tabular}%
\end{table}

\subsection{Efficiency Comparison}
In this subsection, we compare the efficiency of our \method with other baselines, based on two large datasets, i.e., ogbn-arxiv and ogbn-products. To make a fair comparison, we choose the same feature transformation layers for each  method. Besides, we update model parameters with the same iterations in each method, i.e., 500 epochs for ogbn-arxiv and 1,000 epochs for ogbn-products. All the experiments are conducted on the same machine with a NVIDIA RTX A6000 GPU (48 GB memory). For \method, we can update $\vF$ with different frequencies in training, i.e., 1, 2, 3, 4, 5, and ``Full''. ``Full'' means we update both $\vF$ and $\MLP$ in each epoch. For \method-$k$, we only update the $\vF$ for $k$ times during the training procedure. The overall results are shown in Table~\ref{tab:cost}.

\textbf{Training Time.} For the Persistent GNNs including GCN, APPNP and \method-Full,  the training time is longer than other methods that do not need to propagate every epoch. Both APPNP and \method-Full need to propagate ten layers every epoch and GCN needs to propagate three layers. However, the training time of \method-Full is nearly half of APPNP and still less than GCN, which matches our time complexity analysis in Section~\ref{sec:complexity}, as there is no gradient backpropagate through propagation layers. Compared with the One-time GNNs like SGC, \method with only a few update steps, such as \method-5,  can achieve better accuracy with a minor increase in training time.
For example, the whole training time of \method-5 is only 0.92s and 7s longer than SGC, but it has 8.63\% and 11.24 \% relative performance improvements in ogbn-arxiv and ogbn-products datasets, respectively.  Meanwhile, we observe that \method-5 has very similar performance with \method-Full, which suggests that there is no need to do propagation and train the model simultaneously for each epoch. This also suggests that end-to-end training with propagation might not be necessary.

\textbf{Memory Cost.} Compared with the Persistent GNNs, \method requires less memory with no requirement to store the hidden states in the propagation layers. Thus, \method can keep a constant memory even with more propagation layers. Compared with MLP and SGC, \method shows comparable memory. Slightly memory increasing is from the pseudo label matrix as analyzed in Section~\ref{sec:complexity}. \method can be even more efficient with sampling strategies. 

\begin{table*}[!htb]
\caption{Efficiency comparison of different methods.}
\vspace{-0.1in}
\label{tab:cost}
\centering
\resizebox{0.7\linewidth}{!}{%
\begin{tabular}{c|ccc|ccc}
\hline
Dataset      & \multicolumn{3}{c|}{ogbn-arxiv} & \multicolumn{3}{c}{ogbn-products} \\ \hline
Method       & ACC(\%) & Time (s) & Memory (GB) & ACC (\%)  & Time (s) & Memory (GB) \\ \hline
MLP          & 55.68  & 12.01    & 2.68        & 61.17    & 214      & 21.18       \\
SGC          & 66.92  & 12.06    & 2.71        & 74.29    & 215      & 21.85       \\
SIGN         & 71.95  & 24.89    & 4.67        & 80.52    & 492      & 43.17       \\
GCN          & 71.91  & 24.71    & 3.33        & 75.70     & 1,284     & 38.36       \\
APPNP        & 71.61  & 33.70     & 3.20         & 76.62    & 1,913     & 29.15       \\
\method-Full & \textbf{72.76}  & 22.28    & 2.81        & 81.83    & 901      & 24.49       \\
\method-1    & 70.09  & 12.86    & 2.81        &  80.03   & 218      & 24.49       \\
\method-2    & 72.32  & 12.89    & 2.81        &   80.34  & 219      & 24.49       \\
\method-3    & 72.60   & 12.92    & 2.81        & 81.00       & 220      & 24.49       \\
\method-4    & 72.71  & 12.95    & 2.81        & 81.89    & 221      & 24.49       \\
\method-5    & 72.70  & 12.98    & 2.81        & \textbf{82.64}    & 222      & 24.49       \\ \hline
\end{tabular}
}
\end{table*}

In Appendix~\ref{app:efficiency}, we also provide additional experiments to show the efficiency of our \method, i.e. \method can converge faster, need fewer propagation layer, and the memory cost does not increase with more propagation layers.

\subsection{Ablation Study}



\begin{figure}[htbp]
    \vspace{-0.1in}
    \centering
    \subfigure[\centering Cora]{{\includegraphics[width=0.48\linewidth]{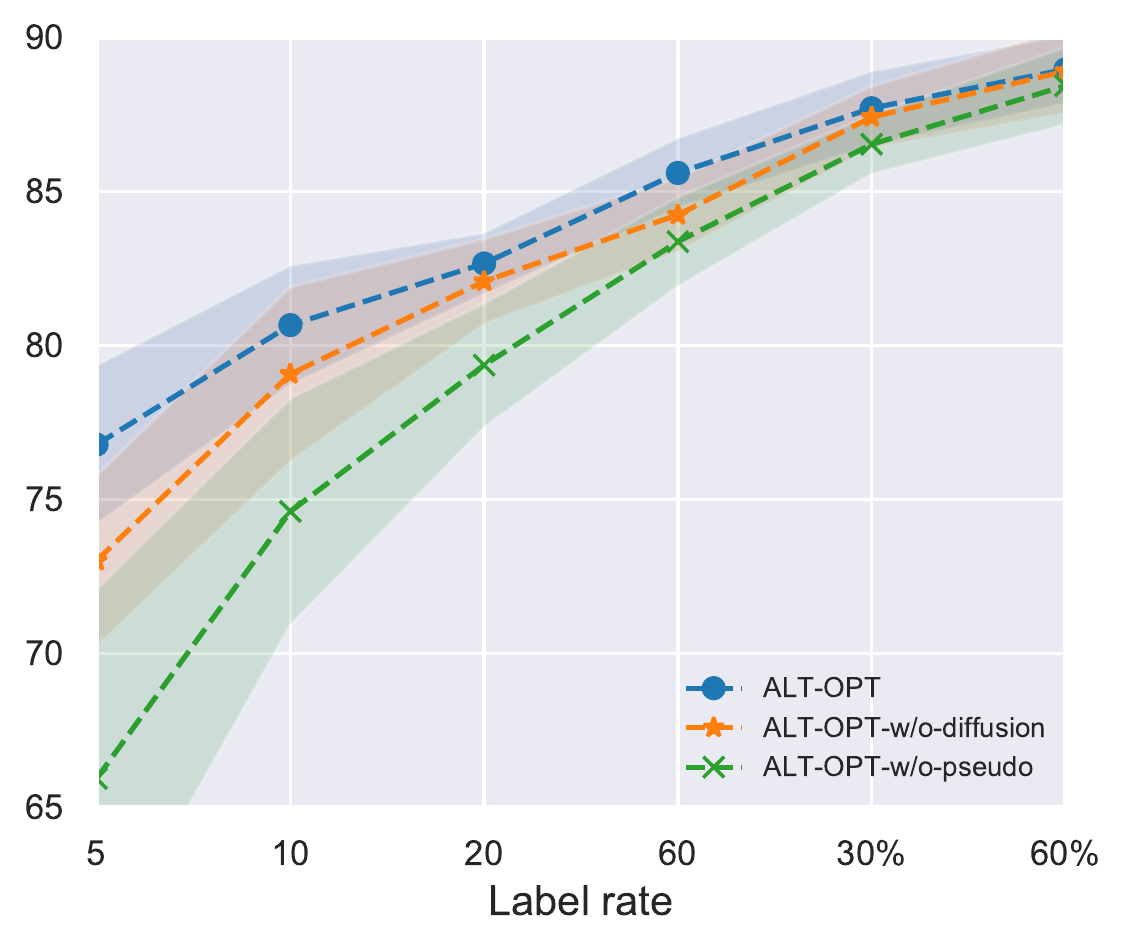}}}
    \hfill
    \subfigure[\centering CiteSeer]{{\includegraphics[width=0.48\linewidth]{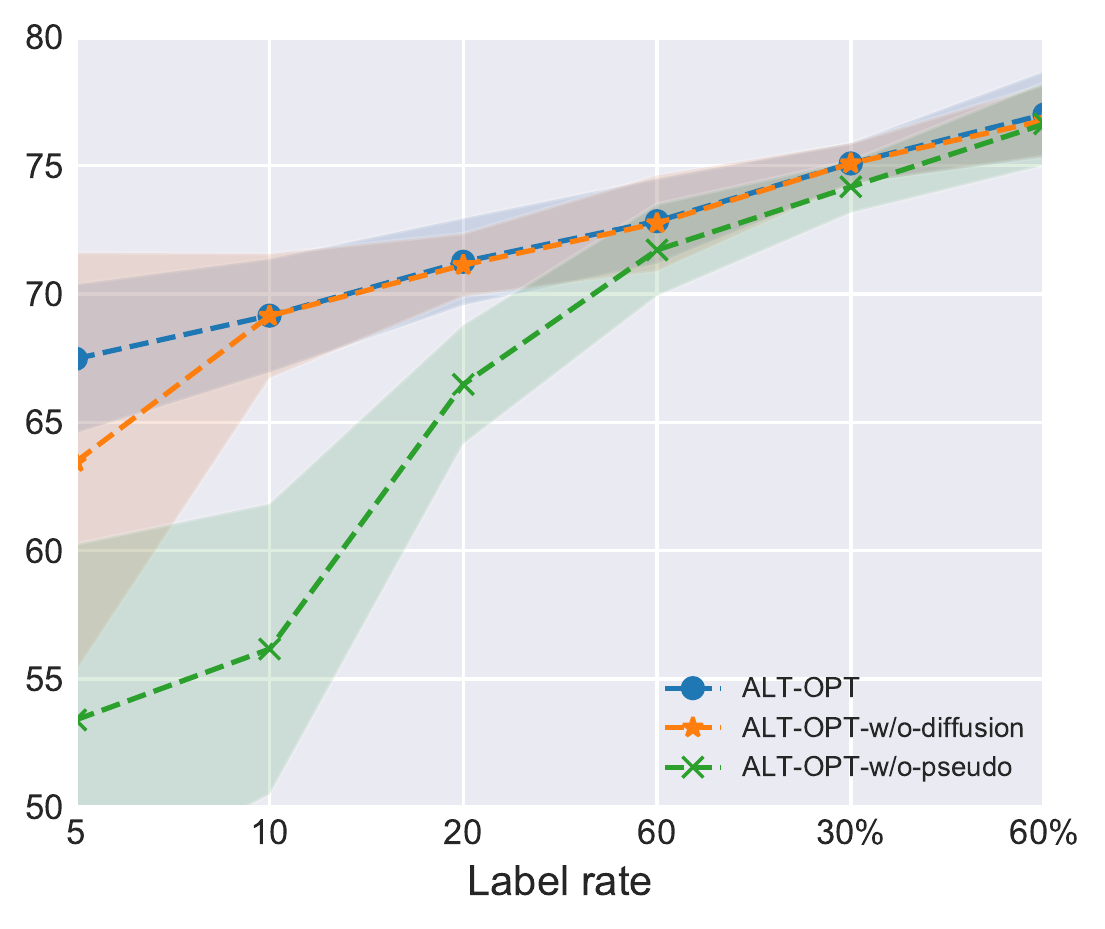} }}
    \vspace{-0.1in}
    \caption{Performance of \method variants.}
    \label{fig:ablation}
    \vspace{-0.1in}
\end{figure}

\noindent \textbf{Feature Diffusion.} It is expected that feature diffusion can improve the accuracy of the $\MLP$ in the pretraining procedure and thus improve the initialization quality of pseudo labels $\vF$. To validate this, we remove the feature diffusion and also use the pseudo labels to train our method, which is called \method-w/o-diffusion. Experiments are conducted on Cora and CiteSeer datasets.
From Figure~\ref{fig:ablation}, we can see that at low label rates, \method is better than \method-w/o-diffusion which means that feature diffusion can boost the model's performance at low labeling rate. As the labeling rate increases, the performance gap becomes small. This shows that feature diffusion is not the key component in our method when the label rate is not very low. 

\begin{figure*}[htb]
\centering
\begin{minipage}[b]{.34\textwidth}
  \vspace{0pt} 
  \centering
  \includegraphics[width=\linewidth, height=1in]{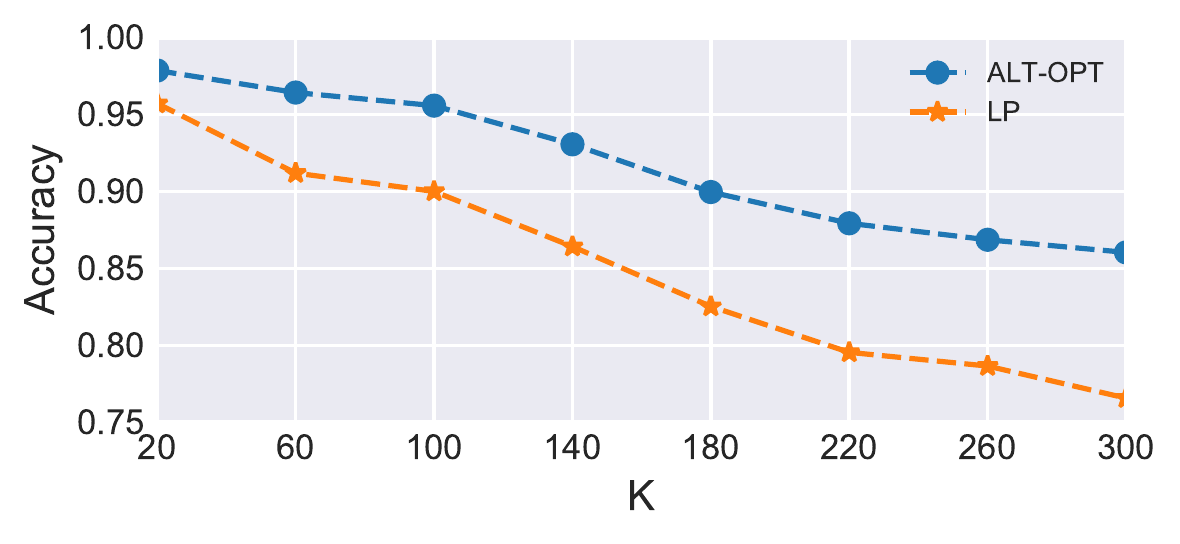}
\end{minipage}%
\begin{minipage}[b]{.64\textwidth}
  \vspace{0pt} 
  \centering
  \includegraphics[width=0.48\linewidth, height=1in]{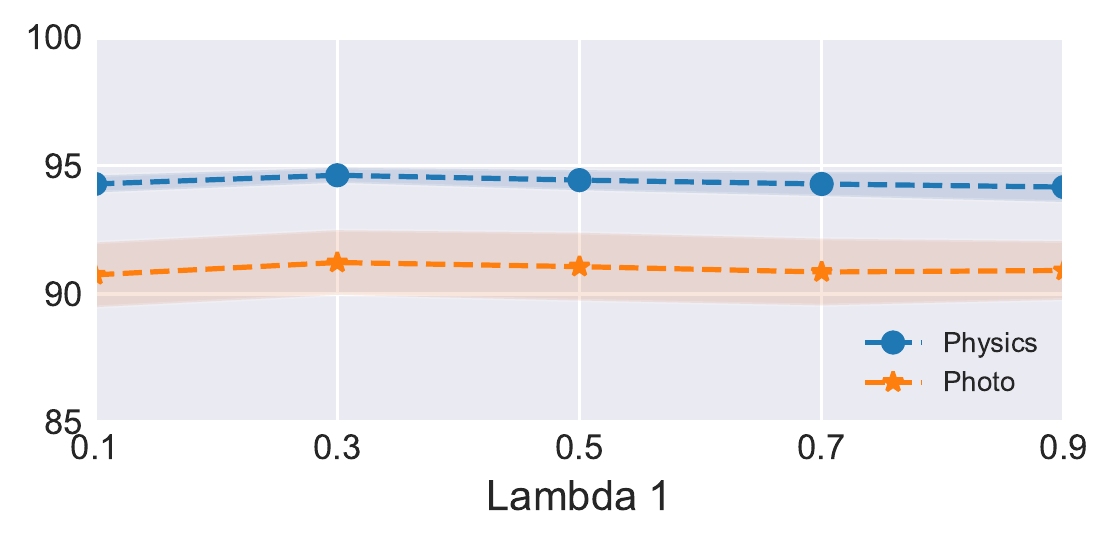}%
  \includegraphics[width=0.48\linewidth, height=1in]{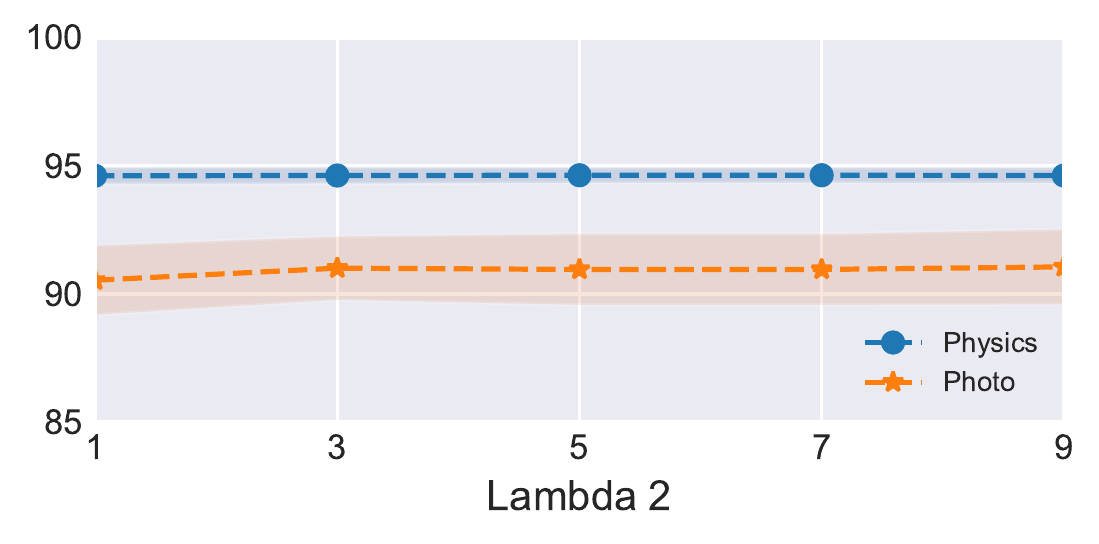}
\end{minipage}
\vspace{-0.1in}
\begin{minipage}[t]{.34\textwidth}
  \captionof{figure}{Top K Accuracy.}
  \label{fig:label}
\end{minipage}%
\begin{minipage}[t]{.64\textwidth}
  \captionof{figure}{Parameter Sensitivity.}
  \label{fig:sen}
\end{minipage}
\vspace{-0.1in}
\end{figure*}

\noindent \textbf{Pseudo Labels.} One of the most important advantages of \method is that we leverage pseudo labels to better train $\MLP$. To study the contribution of pseudo labels in \method, we test the model variant \method-w/o-pseudo which only uses labeled data on Cora and CiteSeer datasets. Compared with \method, Figure \ref{fig:ablation} shows that pseudo labels have a large impact on model performance on both datasets, especially when the label rate is low. 
Moreover, we choose the top $K$ confidence pseudo labels per class after the first update of $\vF$ to verify their accuracy. We adopt the same way to evaluate Label Propagation on Cora dataset with the label rate 20. As shown in Figure \ref{fig:label}, after the first update of $\vF$, the accuracy of the top 180 nodes from each class can be $90\%$. So it is reasonable to use these pseudo labels to train $\MLP$.
Besides, the accuracy of our method at each K is much better than Label propagation, which suggests the effectiveness of the feature-enhanced label propagation update for $\vF$.

\noindent \textbf{Hyperparameters Sensitivity.} We test  the parameter sensitivity of $\lambda_1$ and $\lambda_2$ in Eq~\eqref{eq:2} on Physics and Photo datasets by fixing one with the best parameters and tuning the other. From Figure~\ref{fig:sen}, \method is not very sensitive to these two hyperparameters at the chosen regions.

%% file: 04related.tex
Graph Neural Networks (GNNs) is an effective architecture to represent the graph-structure data, and there are mainly two operators, i.e. feature transformation and  propagation. Based on the order of these two operators, most GNNs can be classified into: Coupled and Decoupled GNNs. Coupled GNNs, such as GCN~\citep{kipf2016semi}, GraphSAGE~\citep{hamilton2017inductive}, and GAT~\citep{velickovic2017graph}, couple feature transformation and propagation in each layer. 
More recently, decoupled GNNs, such as APPNP~\citep{klicpera2018predict}, that decouple the transformation and propagation, are proposed to alleviate the over-smoothness problem~\citep{li2018deeper, oono2019graph}. 
Similar architectures are also utilized in \citep{liu2021elastic, liu2020towards, zhou2021dirichlet}.


In this paper, we categorize GNNs into Persistent GNNs and One-time GNNs. One-time GNNs, such as SGC~\citep{wu2019simplifying}, SIGN~\citep{rossi2020sign} and C\&S~\cite{huang2020combining}, are more efficient than the Persistent GNNs because they only propagate once. PTA \cite{dong2021equivalence} first propagate labels once and then train an MLP, which can be viewed as a One-time GNNs. PPRGo~\citep{bojchevski2020scaling} precomputes the PageRank matrix to avoid multiple steps of propagation, but it still need to do propagation in each epoch, which is also a Persistent GNN. However, our \method does not belong to Coupled/Decoupled or Persistent/One-time GNNs. We propose a new learning paradigm for graph node classification task, which is both efficient and effective.

There are also many sampling methods~\cite{hamilton2017inductive, chen2018fastgcn, zeng2019graphsaint, zou2019layer} that adopt mini-batch training strategies to reduce computation and memory cost by sampling some nodes and edges. Distributed methods~\cite{chiang2019cluster, shao2022distributed} distribute the large graph across multiple servers for parallel training. These works are orthogonal to the contributions in this work and they can be also applied to this work.


\method can be understood as a pseudo-labeling method as discussed in Section~\ref{sec:understanding}. Recent work \cite{iscen2019label} utilizes the label propagation to generate the pseudo labels, which shares some similarity with the proposed ALT-OPT. However, they use the features to generate graph at each iteration, and only the ground truth labels are leveraged for label propagation. For \method, the propagation is a feature enhanced label propagation. We propagate both ground truth labels and features through the given graph. Besides, \method is derived from our proposed framework by an alternating optimization algorithm, which has a different starting point than that of~\citet{iscen2019label}.

%% file: 05conclusion.tex
In this work, we demonstrate that most existing end-to-end GNNs for node classification are solving a bi-level optimization problem. We introduce a new optimization framework for node classification, which can efficiently be optimized with the alternating optimization algorithm. Experimental results validate that \method is both computationally and memory efficient with promising performance on node classification, especially when the labeling rate is low. 

%% file: 06appendix.tex
\section{ \method with Cross Entropy Loss}
\label{app:crossentropy}
In section \ref{sec:altopt}, we instance \method with a Mean Square Error Loss. In this section, we show it can be replaced by a Cross Entropy Loss. By replacing the first part of Eq~\eqref{eq:2} to be a cross entropy between $\MLP$ and $\vF$, the formulation becomes:
\begin{align} 
   \mathcal{L} =  \lambda_1 CE\Big(\MLP(\vX), \vF\Big) + \tr(\vF^\top \tL \vF) + \lambda_2 \|\vF_L-\vY_L\|_F^2
\end{align}
where $CE(\cdot, \cdot)$ is the cross entropy function. Adopting the same gradient decent method, the update rule for $\vF$ becomes:
\begin{align}
\vF^{k+1}_L&=\vF^{k}_L-\eta_L\left(-\lambda_{1} \log \MLP(\vX_L)+2 (\tL \vF^{k})_L+2\lambda_{2}(\vF^{k}_L-\vY_L)\right) \nonumber \\
&=\left(1-2 \eta_L\left(1+\lambda_{2}\right)\right) \vF^{k}_L+\eta_L \lambda_{1} \log \MLP(X)_L+2 \eta_L \tA \vF^{k}_L+2 \eta_L \lambda_{2} \vY_L, \\
\vF^{k+1}_U&=\vF^{k}_U-\eta_U\left(-\lambda_{1} \log \MLP(\vX_U)+2 (\tL \vF^{k})_U \right) \nonumber \\
&=\left(1-2 \eta_U \right) \vF^{k}_U+\eta_U \lambda_{1} \log \MLP(\vX_U)+2 \eta_U (\tA \vF^{k})_U
\end{align}
where the $\MLP$ is the output after the Softmax function and the step size can be set as $\eta_L = \eta_U = \frac{1}{2(1+\lambda_2)}$. 
Therefore, the update rule of $\vF$ is:
\begin{align}
    \vF^{k+1}_L &= \frac{\lambda_{1}}{2\left(1+\lambda_{2}\right)} \log \MLP(\vX_L)+\frac{1}{1+\lambda_{2}} (\tA \vF^{k})_L+\frac{\lambda_{2}}{1+\lambda_{2}} \vY_L, \\
    \vF^{k+1}_U &= \frac{\lambda_{1}}{2(1+\lambda2)} \log \MLP(\vX_U)+ \frac{1}{1+\lambda_2} (\tA \vF^{k})_U + \frac{\lambda_2}{1+\lambda_2}\vF^k_U
\end{align}
Then we can consider the hidden variable $\vF$ as pseudo label and update the parameters of $\MLP$ based on the cross entropy loss $CE(\MLP, \vF)$. In practice, there are some situations that the cross entropy have a better performance than the original mean square error loss. Using two different losses would give similar results in most cases, and we report the best one.

\section{Understandings of \method}
\subsection{Comparison between updating F and Label Propagation}
\label{app:lp}

Label Propagation (LP) \citep{zhou2003learning} is a well-known graph semi-supervised learning method based on the label smoothing assumption that connected nodes are likely to have the same label. The label propagation can be written as solving an the following optimization problem:
\begin{align} 
\label{eq:lpo}
\begin{aligned}
    \mathcal{Q}(\vP) &=\frac{1}{2}\sum_{i, j=1}^{n} A_{ij}\left\|\frac{1}{\sqrt{D_{ii}}} \vP_{i}-\frac{1}{\sqrt{D_{jj}}} \vP_{j}\right\|^{2}+\mu \sum_{i=1}^{n}\left\|\vP_{i}-\vY_{i}\right\|^{2} \\
    & = tr(\vP^\top \tL \vP) + \mu \|\vP-\vY\|_F^{2}
\end{aligned}
\end{align}

We can use an iteration algorithm to solve  Eq.~\ref{eq:lpo}. The $k$-th iteration process of LP is as follows:
 \begin{align}
 \label{eq:app.lp}
     \vP(k)= (1-\alpha) \tA \vP(k-1)+ \alpha \vY,
 \end{align}
 where $\tA$ is the normalized graph Laplacian matrix, $\vY$ is the label matrix, $\vP(0) = \vY$, and $\alpha \in(0, 1)$ is a hyperparamter. 

If we iterate Eq.~\ref{eq:app.lp} for one time, the $\vP(k)$ becomes:
 \begin{align}
 \begin{aligned}
 \label{eq:app.lp1}
     \vP(k) &= (1-\alpha) \tA \left((1-\alpha) \tA \vP(k-2)+ \alpha \vY\right)+ \alpha \vY, \\
     & = (1-\alpha)^2 \tA^2 \vP(k-2) + \alpha (1-\alpha) \tA \vY + \alpha \vY
\end{aligned}
 \end{align}

By iterating $K$ times, we can get:
\begin{align}
\label{eq:lpo1}
\begin{aligned}
    \vP(K) &= (1-\alpha)^K {\tA}^K \vY +\alpha \sum_{k=0}^{K-1}(1-\alpha)^k \tA^k \vY \\
    & = \left((1-\alpha)^K {\tA}^K  +\alpha \sum_{k=0}^{K-1}(1-\alpha)^k \tA^k  \right) \vY\\
    & = \bar{\vA} \vY
\end{aligned}
\end{align}

Let $\bar{\vA} = (1-\alpha)^K {\tA}^K  +\alpha \sum_{k=0}^{K-1}(1-\alpha)^k \tA^k $, then the $K$ step LP can be represented as $LP(\vY, \alpha) = \bar{\vA} \vY$.
For our \method, our optimization problem is:
\begin{align*}
   \mathcal{L} =  \lambda_1\|\MLP(\vX)-\vF\|_F^2 + \tr(\vF^\top \tL \vF) + \lambda_2 \|\vF-\vY\|_F^2,
\end{align*}
which can also be written as 
\begin{align}
\label{eq:alto}
\vF(k) &= \frac{1}{\lambda_1+\lambda_2+1} \tA\vF(k-1) + \frac{\lambda_1}{\lambda_1+\lambda_2+1} \MLP(\vX) + \frac{\lambda_2}{\lambda_1+\lambda_2+1} \vY.
\end{align}
By iterating Eq.~\ref{eq:alto}, it becomes:
\begin{align}
\label{eq:alto1}
\begin{aligned}
    \vF(k) = \left((\frac{1}{\lambda_1+\lambda_2+1})^K {\tA}^K  +\frac{\lambda_1 + \lambda_2}{\lambda_1+\lambda_2+1}  \sum_{k=0}^{K-1}(\frac{1}{\lambda_1+\lambda_2+1})^k \tA^k  \right) \left(\frac{\lambda_1}{\lambda_1+\lambda_2+1} \MLP(\vX) + \frac{\lambda_2}{\lambda_1+\lambda_2+1} \vY \right)
\end{aligned}
\end{align}

Let $\alpha=\frac{\lambda_1 + \lambda_2}{\lambda_1+\lambda_2+1}, \beta = \frac{\lambda_2}{\lambda_1+\lambda_2+1}$,  Eq.~\ref{eq:alto1} can be written as:
\begin{align}
\label{eq:alto2}
    \vF(K) = \bar{\vA} \left((\alpha-\beta)\MLP(\vX) + \beta \vY \right).
\end{align}

Compared Eq.~\ref{eq:lpo} with Eq.~\ref{eq:alto2}, \method not only propagates the ground truth labels, but ``feature" labels $\MLP(\vX)$ generated by features. Thus, updating $\vF$ takes advantage of all node features, graph structure and labels, while LP only leverages graph structure and labels. Thus, updating $\vF$ of \method can be seen as a feature-enhanced label propagation.

\subsection{Comparison between updating $\Theta$ and pseudo-labeling methods}
\label{app:pseudo}
Pseudo-labeling \citep{lee2013pseudo, arazo2020pseudo} is a popular method in semi-supervised learning that uses a small set of labeled data along with a large amount of unlabeled data to improve model performance. It usually generates pseudo labels for the unlabeled data and trains the deep models using both the true labels and pseudo labels with different weights. From this perspective, \method{} uses the pseudo labels $\vF$ to train $\MLP$.

Let $\boldsymbol{f}_\theta(\boldsymbol{X})=\MLP(X)$, the loss function for training \method can be written as the following based on Eq.~\ref{eq:alto2}:
\begin{align}
\label{eq:loss}
L_{\method} & =\ell\left(\boldsymbol{f}_\theta(\boldsymbol{X}), \overline{\boldsymbol{A}} ((\alpha - \beta) \boldsymbol{f}_\theta(\boldsymbol{X})^\prime + \beta \boldsymbol{Y}) \right),
\end{align}
where $\boldsymbol{f}_\theta(\boldsymbol{X})^\prime$ is the output from previous layer, and there is no gradient information.

If we choose Cross Entropy loss, then Eq.~\ref{eq:loss} becomes:
\begin{align}
\label{eq:loss1}
\begin{aligned}
    L_{\method} &= - \sum_{i, j \in \cV} \sum_{k\in \cC} \bar{a}_{i j} ((\alpha - \beta) f_\theta(\vX)_{j, k}^\prime + \beta \vY_{jk}) log f_\theta(\vX)_{i, k} \\
    & = -(\alpha-\beta) \sum_{i, j \in \cV} \sum_{k\in \cC} \bar{a}_{i j} f_\theta(\vX)_{j, k}^\prime log f_\theta(\vX)_{i, k} - \beta \sum_{i, j \in \cV} \sum_{k\in \cC}\vY_{jk}) log f_\theta(\vX)_{i, k} \\
    &= -(\alpha-\beta) \sum_{i, j \in \cV} \sum_{k\in \cC} \bar{a}_{i j} f_\theta(\vX)_{j, k}^\prime log f_\theta(\vX)_{i, k} - \beta \sum_{i, j \in \cV} \sum_{k\in \cC} \bar{a}_{i j} \vY_{jk} log f_\theta(\vX)_{i, k} \\
    &=  -(\alpha-\beta) \sum_{i, j \in \cV} \sum_{k\in \cC} \bar{a}_{i j} f_\theta(\vX)_{j, k}^\prime log f_\theta(\vX)_{i, k} - \beta \sum_{i\in \cV, j \in \cV_L} \vY_{j h(j)} log f_\theta(\vX)_{i, h(j)} \\
    &= (\alpha - \beta) \sum_{i, j \in \cV} \bar{a}_{i j} CE(f_i, f_j^\prime) + \beta \sum_{i\in \cV, j \in \cV_L} \bar{a}_{i j} CE(f_i, y_j),
\end{aligned}
\end{align}
where $h(j)$ is the label of labeled node $j$, thus $\vY_{j, k} = 1$ if $k=h(j)$ else $\vY_{j,k} = 0$. From Eq.~\ref{eq:loss1}, we can find that \method use both truth label $y_i$ and pseudo label $f_j^\prime$ to train the $\MLP$.

\section{Incorporating Normalization and Pseudo-label Reweighting into a Unified Framework}
\label{app:incoporate}

In our \method implementation, we utilize normalization to map the feature into a label space and a pseudo-label reweighting operator to select high-confidence nodes for training $\MLP$. These operations cannot be directly derived from Eq.~\ref{eq:2}. However, by slightly modifying our approach, we can integrate these two operators into a unified framework. Specifically, we use the softmax function to map $\mathbf{F}$ into the label space and a diagonal weight matrix to reweight pseudo-labels. Let $softmax(\mathbf{F}) = S(\mathbf{F})$. We can then modify the optimization target as follows:

\begin{align}
    L =  \lambda_1 \mathbf{W} CE( MLP(\mathbf{X}), S(\mathbf{F})) + \operatorname{tr}\left(\mathbf{F}^{\top} \tilde{\mathbf{L}} \mathbf{F}\right)+\lambda_2 CE(Y, S(\mathbf{F})),
\end{align}
where CE is the Cross-Entropy function, the weight matrix $\mathbf{W}$ is a diagonal matrix, 
$w_q = w_q^\prime \mathbbm{1} \left(w^{\prime}_q > \tau \right)$, $w^\prime_q = Entropy(softmax(\mathbf{F}_q)))$, $\mathbbm{1}$ is the indicator function, and $\tau$ is a threshold.
The reason we adopt the Cross-Entropy function is that it has a neat derivative with the Softmax operator, i.e. $\frac{\partial{CE(\mathbf{Y}, S(\mathbf{F}))}}{\partial{\mathbf{F}}} = S(\mathbf{F}) - \mathbf{Y}$. 

Therefore, the gradient of $\mathcal{L}$ with respect to $\mathbf{F}$ is:
\begin{align}
    \frac{\partial{L}}{\partial{F}} = \lambda_1 \mathbf{W}  (S(\mathbf{F}) - MLP(\mathbf{X})) + 2  \mathbf{L} \mathbf{F} + \lambda_2 (S(\mathbf{F}) - Y).
\end{align}
This gradient closely resembles the original one with the MSE loss, making the optimization target quite tidy and similar to the original one.
This problem can be resolved using the same alternating optimization method as before. We term the new method as ALT-OPT-N. To evaluate our proposed method, we chose three representative datasets, namely, PubMed, Coauthor CS, and ogbn-arxiv. The results are as follows:

\begin{table}[htb]
\centering
\caption{Different implementations of ALT-OPT}
\begin{tabular}{c|ccc}
\hline
Dataset     & PubMed       & CS           & ogbn-arxiv \\ \hline
ALT-OPT     & 79.16 ± 2.26 & 92.77 ± 0.50 & 72.60      \\
ALT-OPT-N & 78.66 ± 2.00 & 92.69 ± 0.28 & 72.58      \\ \hline
\end{tabular}
\end{table}

The results show that ALT-OPT-N achieves performance similar to ALT-OPT, with all operators being derivable from ALT-OPT-N. This underlines the efficacy of our proposed unified framework for incorporating normalization and pseudo-label reweighting.

\section{Algorithm of \method}
\label{app:algorithm}
In this section, we provide the algorithm~\ref{alg:alg} of \method.

We first initialize the pseudo label $\vF$ as label matrix $\vY$, and then preprocess data by feature diffusion. Afterward, we pre-train the MLP on the labeled data for a few epochs. Then, we update $\vF$ and $\MLP$ alternatively and iteratively. 

\begin{algorithm}[htb]
\caption{Algorithm of \method}
\label{alg:alg}
\begin{algorithmic}
\STATE {\bfseries Input:} Adjacent matrix $\vA$, Features $\vX$, Labels $\vY$, Hyperparamters $\lambda_1, \lambda_2, \alpha$, pseudo label number $m$, pretraining steps $s$, MLP update times $t$
\STATE {\bfseries Output}: Pseudo Label $\vF$, MLP parameter $\Theta$ \\
\STATE Initialize $\vF = \vY$
\STATE $\vX^\prime = LP(\vX, \alpha)$
\FOR{$i=1$ {\bfseries to} $s$}
   \STATE Pretrain $\MLP$ by $\mathcal{L}= \sum_{i\in L} \ell(\MLP(x^{\prime}_i; \Theta), \vY_i)$
   \ENDFOR
   \REPEAT
   \STATE Update $\vF$ based on Eq.~\eqref{eq:F_L} and Eq.~\eqref{eq:F_U}
    \STATE Normalize $\vF$ based on $\vF_{ij} = \frac{\exp \left(\vF_{ij} / \tau\right)}{\sum_{k=1}^{C} \exp \left(\vF_{ik} / \tau\right)}$ 
    \STATE Select $m$ top unlabeled nodes $U_t$ per class by  $w_i = 1 - \frac{H(\vF_i)}{log(C)}$
    \FOR{$i=1$ {\bfseries to} $t$}
    \STATE Update $\Theta$ by minimizing $\mathcal{L}_{MLP}(X^{\prime}, F; \Theta)=\sum_{i\in L\cup U_t} w_i \ell(\MLP(x^{\prime}_i; \Theta), \vF_i)$ 
    \ENDFOR
    \UNTIL{Model Converge}
\end{algorithmic}
\end{algorithm}

\section{Datasets Statistics}
\label{app:dataset}

In the experiments, the data statistics used in Section \ref{sec:exp} are summarized in Table~\ref{tab:dataset}. For Cora, CiteSeer and PubMed dataset, we adopt different label rates, i.e., 5, 10, 20, 60, 30\% and 60\% labeled nodes per class, to get a more comprehensive comparison.  For label rates 5, 10, 20, and 60, we use 500 nodes for validation and 1000 nodes for test. For label rates  30\% and 60\%, we use half of the rest nodes for validation and the remaining half for test. For each labeling rate, we adopt 10 random splits for each dataset. For other datasets, we follow the original data split.

\begin{table}[htb]
\centering
\caption{Dataset Statistics.}
\label{tab:dataset}
\begin{tabular}{c|cccc}
\hline
Dataset       & Nodes     & Edges      & Features & Classes \\ \hline
Cora          & 2,708     & 5,278      & 1,433    & 7       \\
CiteSeer      & 3,327     & 4,552      & 3,703    & 6       \\
PubMed        & 19,717    & 44,324     & 500      & 3       \\
Coauthor CS            & 18,333    & 81,894     & 6,805    & 15      \\
Coauthor Physics       & 34,493    & 247,962    & 8,415    & 5       \\
Amazon Computer      & 13,381    & 245,778    & 767      & 10      \\
Amazon Photo         & 7,487     & 119,043    & 745      & 8       \\
Flickr        & 89,250    & 899,756    & 500      & 7       \\
Reddit        & 232,965   & 11,606,919 & 602      & 41      \\
Ogbn-Arxiv    & 169,343   & 1,166,243  & 128      & 40      \\
Ogbn-Products & 2,449,029 & 61,859,140 & 100      & 47      \\ \hline
\end{tabular}
\end{table}

\section{Parameters Setting}
\label{app:parameter}
In this section, we describe in detail the search space for parameters of different experiments.
\subsection{Transductive Setting}
For all deep models, we use 3 transformation layers with 256 hidden units for OGB datasets, and 2 transformation layers with 64 hidden units for other datasets. For all methods, the following hyperparameters are tuned based on the loss and validation accuracy from the following search space: 
\begin{itemize}
\vspace{-0.1in}
\setlength\itemsep{-0.2em}
    \item Learning Rate: \{0.01, 0.05\}
    \item Dropout Rate: \{0, 0.5, 0.8\}
    \item Weight Decay:  \{ 5e-4, 5e-5, 0\}
    \item Hyperparamters between 0 and 1: step size 0.1
\end{itemize}
The propagation layers for APPNP and C\&S is tuned from \{5, 10\} and \{10, 20, 50\}, respectively. 

For \method, the $\lambda_1$ and $\lambda_2$ are tuned from \{0.1, 0.3, 0.5, 0.7, 1\} and \{1, 3, 5, 7, 10\}, respectively; 10 propagation layers; pretraining steps $s=100$; $\tau =0.1$; pseudo label numbers per class $m$ are choose from \{100, 200, 500, 5000\} based on the size of graphs; The training epochs e is set to 1,000 for ogbn-products dataset and 500 for all other datasets same as other models. Then, a propagation times k is chosen (for example 5). Afterwards, we evenly split the training epochs $e$ to $k$ parts. For the first $e/k$ epochs, we train the $\MLP$, and then do propagation once. For the next $e/k$ epochs we train the $MLP$ and then propagate once, and so on. 

The Adam optimizer\citep{kingma2014adam} is used in all experiments.

\subsection{Inductive Setting}
For the inductive node classification task, we follow the data process as previous work \cite{zeng2019graphsaint}. Specifically, we first filter the training graph that only contains labeled node for training, and the entire graph are used for inference. For all models, we use 3 transformation layers with 256 hidden units for Reddit dataset, and 2 layers with 64 hidden units for Flickr dataset.  Besides, we adopt the most hyper-parameters search space for all baselines. The propagation step K for APPNP and C\& is tuned from \{2, 3, 5, 10\} and \{10, 20, 50\}, respectively. For C\&S, both correct and smooth hyper-parameters are choose from [0,1] with granularity of 0.1. The $\lambda_1$ and $\lambda_2$ in \method are tuned with granularity of 0.1 in range [0, 1], and 1 in [1, 10], respectively.

\section{Transductive Node Classification Results}
\label{app:transductive}
For the transductive semi-supervised node classification task, we choose nine common used datasets including three citation datasets, i.e., Cora, Citeseer and Pubmed \citep{sen2008collective}, two coauthors datasets, i.e., CS and Physics, two Amazon datasets, i.e., Computers and Photo \citep{shchur2018pitfalls}, and two OGB datasets, i.e., ogbn-arxiv and ogbn-products \citep{hu2020open}.  

We compare the proposed \method with three groups of methods: (i) Persistent GNNs, i.e.,  GCN~\citep{kipf2016semi}, GAT~\citep{velivckovic2017graph} and APPNP~\citep{klicpera2018predict}; (ii) One-time GNNs, i.e., SGC~\citep{wu2019simplifying}, SIGN~\cite{rossi2020sign}, and C\&S \citep{huang2020combining}; and (iii) Non-GNN methods including MLP and Label Propagation\citep{zhou2003learning}. The overall performance are shown in Table~\ref{tab:all1}.

\begin{table}[htb]
\centering
\caption{The overall results of the transductive node classification task.}
\label{tab:all1}
\resizebox{\columnwidth}{!}{%
\begin{tabular}{cc|cc|ccc|ccc|c}
\hline
\multicolumn{2}{c|}{Method}                     & \multicolumn{2}{c|}{Non-GNN}                             & \multicolumn{3}{c|}{Persistent GNNs}                         & \multicolumn{3}{c|}{One-time GNNs}               & Ours                      \\ \hline
\multicolumn{1}{c|}{Dataset}       & Label Rate & \multicolumn{1}{c|}{LP}               & MLP              & GCN              & GAT              & APPNP                     & SGC              & SIGN         & C\&S             & \method                   \\ \hline
\multicolumn{1}{c|}{Cora}          & 5          & \multicolumn{1}{c|}{57.60 $\pm$ 5.71} & 42.34 $\pm$ 3.31 & 70.68 $\pm$ 2.17 & 72.97 $\pm$ 2.23 & 75.86 $\pm$ 2.34          & 70.06 $\pm$ 1.95 & 69.81 $\pm$ 3.13 & 56.52 $\pm$ 5.53 & \textbf{76.78 $\pm$ 2.56} \\
\multicolumn{1}{c|}{}              & 10         & \multicolumn{1}{c|}{63.76 $\pm$ 3.60} & 51.34 $\pm$ 3.37 & 76.50 $\pm$ 1.42 & 78.03 $\pm$ 1.17 & 80.29 $\pm$ 1.00          & 76.28 $\pm$ 1.22 & 76.25 $\pm$ 1.26 & 71.04 $\pm$ 3.30 & \textbf{80.66 $\pm$ 1.92} \\
\multicolumn{1}{c|}{}              & 20         & \multicolumn{1}{c|}{67.87 $\pm$ 1.43} & 59.23 $\pm$ 2.52 & 79.41 $\pm$ 1.30 & 81.39 $\pm$ 1.41 & 82.34 $\pm$ 0.67          & 80.30 $\pm$ 1.72 & 79.71 $\pm$ 1.11 & 77.96 $\pm$ 2.13 & \textbf{82.66 $\pm$ 0.98} \\
\multicolumn{1}{c|}{}              & 60         & \multicolumn{1}{c|}{73.92 $\pm$ 1.25} & 68.35 $\pm$ 2.08 & 84.30 $\pm$ 1.44 & 85.11 $\pm$ 1.10 & 85.49 $\pm$ 1.25          & 84.17 $\pm$ 1.39 & 84.16 $\pm$ 1.18 & 82.21 $\pm$ 1.45 & \textbf{85.60 $\pm$ 1.12} \\
\multicolumn{1}{c|}{}              & 30\%       & \multicolumn{1}{c|}{82.26 $\pm$ 1.89} & 73.26 $\pm$ 1.38 & 86.87 $\pm$ 1.35 & 87.24 $\pm$ 1.19 & \textbf{87.77 $\pm$ 1.13} & 86.97 $\pm$ 0.90 & 87.17 $\pm$ 1.28 & 87.60 $\pm$ 1.12 & 87.70 $\pm$ 1.19          \\
\multicolumn{1}{c|}{}              & 60\%       & \multicolumn{1}{c|}{86.05 $\pm$ 1.35} & 76.49 $\pm$ 1.13 & 88.60 $\pm$ 1.19 & 88.68 $\pm$ 1.13 & 88.49 $\pm$ 1.28          & 88.60 $\pm$ 1.38 & 88.21 $\pm$ 1.11 & 88.68 $\pm$ 1.39 & \textbf{88.96 $\pm$ 1.10} \\ \hline
\multicolumn{1}{c|}{CiteSeer}      & 5          & \multicolumn{1}{c|}{39.06 $\pm$ 3.53} & 41.05 $\pm$ 2.84 & 61.27 $\pm$ 3.85 & 62.60 $\pm$ 3.34 & 63.92 $\pm$ 3.39          & 60.21 $\pm$ 3.48 & 57.44 $\pm$ 3.71 & 50.39 $\pm$ 4.70 & \textbf{67.48 $\pm$ 2.90} \\
\multicolumn{1}{c|}{}              & 10         & \multicolumn{1}{c|}{42.29 $\pm$ 3.26} & 47.99 $\pm$ 2.71 & 66.28 $\pm$ 2.14 & 66.81 $\pm$ 2.10 & 67.57 $\pm$ 2.05          & 65.23 $\pm$ 2.36 & 63.87 $\pm$ 3.09 & 58.96 $\pm$ 2.75 & \textbf{69.39 $\pm$ 2.59} \\
\multicolumn{1}{c|}{}              & 20         & \multicolumn{1}{c|}{46.15 $\pm$ 2.31} & 56.96 $\pm$ 1.80 & 69.60 $\pm$ 1.67 & 69.66 $\pm$ 1.47 & 70.85 $\pm$ 1.45          & 68.82 $\pm$ 2.11 & 68.60 $\pm$ 1.94 & 65.85 $\pm$ 2.74 & \textbf{71.26 $\pm$ 1.69} \\
\multicolumn{1}{c|}{}              & 60         & \multicolumn{1}{c|}{52.76 $\pm$ 1.14} & 66.37 $\pm$ 1.56 & 72.52 $\pm$ 1.74 & 73.10 $\pm$ 1.20 & \textbf{73.50 $\pm$ 1.54} & 71.43 $\pm$ 1.26 & 72.63 $\pm$ 1.39 & 71.21 $\pm$ 1.79 & 72.84 $\pm$ 1.65          \\
\multicolumn{1}{c|}{}              & 30\%       & \multicolumn{1}{c|}{62.75 $\pm$ 1.30} & 70.37 $\pm$ 1.00 & 75.20 $\pm$ 0.85 & 75.01 $\pm$ 0.99 & \textbf{75.71 $\pm$ 0.71} & 75.09 $\pm$ 1.01 & 74.44 $\pm$ 0.83 & 74.65 $\pm$ 0.95 & 75.09 $\pm$ 0.79          \\
\multicolumn{1}{c|}{}              & 60\%       & \multicolumn{1}{c|}{69.39 $\pm$ 2.01} & 73.15 $\pm$ 1.36 & 76.88 $\pm$ 1.78 & 76.70 $\pm$ 1.81 & \textbf{77.42 $\pm$ 1.47} & 76.66 $\pm$ 1.59 & 76.41 $\pm$ 1.96 & 76.34 $\pm$ 1.37 & 77.00 $\pm$ 1.67          \\ \hline
\multicolumn{1}{c|}{Pubmed}        & 5          & \multicolumn{1}{c|}{65.52 $\pm$ 6.42} & 58.48 $\pm$ 4.06 & 69.76 $\pm$ 6.46 & 70.42 $\pm$ 5.36 & 72.68 $\pm$ 5.68          & 68.55 $\pm$ 6.88 & 66.52 $\pm$ 6.15 & 65.3 $\pm$ 6.02  & \textbf{73.51 $\pm$ 4.80} \\
\multicolumn{1}{c|}{}              & 10         & \multicolumn{1}{c|}{68.39 $\pm$ 4.88} & 65.36 $\pm$ 2.08 & 72.79 $\pm$ 3.58 & 73.35 $\pm$ 3.83 & 75.53 $\pm$ 3.85          & 72.80 $\pm$ 3.55 & 71.32 $\pm$ 3.70 & 72.51 $\pm$ 3.75 & \textbf{75.55 $\pm$ 5.09} \\
\multicolumn{1}{c|}{}              & 20         & \multicolumn{1}{c|}{71.88 $\pm$ 1.72} & 69.07 $\pm$ 2.10 & 77.43 $\pm$ 1.93 & 77.43 $\pm$ 2.66 & 78.93 $\pm$ 2.11          & 76.48 $\pm$ 2.84 & 76.39 $\pm$ 2.65 & 75.34 $\pm$ 2.49 & \textbf{79.16 $\pm$ 2.26} \\
\multicolumn{1}{c|}{}              & 60         & \multicolumn{1}{c|}{75.79 $\pm$ 1.54} & 76.20 $\pm$ 1.48 & 82.00 $\pm$ 1.62 & 81.40 $\pm$ 1.40 & \textbf{82.55 $\pm$ 1.47} & 80.34 $\pm$ 1.61 & 81.75 $\pm$ 1.55 & 80.63 $\pm$ 1.49 & 82.53 $\pm$ 1.76          \\
\multicolumn{1}{c|}{}              & 30\%       & \multicolumn{1}{c|}{82.51 $\pm$ 0.34} & 85.92 $\pm$ 0.25 & 88.07 $\pm$ 0.29 & 86.51 $\pm$ 0.41 & 87.56 $\pm$ 0.39          & 86.23 $\pm$ 0.43 & 89.09 $\pm$ 0.33 & 88.44 $\pm$ 0.40 & 88.24 $\pm$ 0.36          \\
\multicolumn{1}{c|}{}              & 60\%       & \multicolumn{1}{c|}{83.38 $\pm$ 0.64} & 86.14 $\pm$ 0.64 & 88.48 $\pm$ 0.46 & 86.52 $\pm$ 0.56 & 87.56 $\pm$ 0.52          & 86.63 $\pm$ 0.38 & 89.55 $\pm$ 0.56 & 88.53 $\pm$ 0.56 & \textbf{88.83 $\pm$ 0.55} \\ \hline
\multicolumn{1}{c|}{CS}            &    20        & \multicolumn{1}{c|}{77.45 $\pm$ 1.80}     & 88.12 $\pm$ 0.78 & 91.73 $\pm$ 0.49     & 90.96 $\pm$ 0.46     & 92.38 $\pm$ 0.38              & 90.32 $\pm$ 0.99     & 92.02 $\pm$ 0.41 & 92.41 $\pm$ 0.44     & \textbf{92.77} $\pm$ 0.50              \\ \hline
\multicolumn{1}{c|}{Physics}       &      20      & \multicolumn{1}{c|}{86.70 $\pm$ 1.03}     & 88.30 $\pm$ 1.59     & 93.29 $\pm$ 0.80     & 92.81 $\pm$ 1.03     & 93.49 $\pm$ 0.67              & 93.23 $\pm$ 0.59     & 93.03 $\pm$ 1.15 & 93.23 $\pm$ 0.55     & \textbf{94.63 $\pm$ 0.31}              \\ \hline
\multicolumn{1}{c|}{Computers}     &     20       & \multicolumn{1}{c|}{72.44 $\pm$ 2.87}     & 60.66 $\pm$ 2.98     & 79.17 $\pm$ 1.92     & 78.38 $\pm$ 2.27     & 79.07 $\pm$ 2.34              & 73.00 $\pm$ 2.0      & 73.04 $\pm$1.15  & 73.25$\pm$ 2.09      & \textbf{79.12 $\pm$ 2.50}              \\ \hline
\multicolumn{1}{c|}{Photo}         &     20       & \multicolumn{1}{c|}{81.58 $\pm$ 4.69}     & 75.33 $\pm$ 1.91     & 89.94 $\pm$ 1.22     & 89.24 $\pm$ 1.42     & 90.87 $\pm$ 1.14              & 83.50 $\pm$ 2.9      & 86.11 $\pm$ 0.66 & 84.87 $\pm$ 1.04     & \textbf{91.23} $\pm$ 1.26              \\ \hline
\multicolumn{1}{c|}{ogbn-arxiv}    & 54\%       & \multicolumn{1}{c|}{68.14 $\pm$ 0.00} & 55.68 $\pm$ 0.11 & 71.91 $\pm$ 0.15 & 71.92 $\pm$ 0.17 & 71.61 $\pm$ 0.30          & 68.74 $\pm$ 0.12 & 71.95 $\pm$ 0.11 & 71.03 $\pm$ 0.15 & \textbf{72.76 $\pm$ 0.17}     \\ \hline
\multicolumn{1}{c|}{ogbn-products} & 8\%        & \multicolumn{1}{c|}{74.08 $\pm$ 0.00}     & 61.17 $\pm$ 0.20     & 75.70 $\pm$ 0.19     & OOM              & 76.62 $\pm$ 0.13              & 73.15 $\pm$ 0.12     & 80.52$\pm$0.16   &  77.11 $\pm$ 0.06     & \textbf{82.64$\pm$ 0.21}      \\ \hline
\end{tabular}
}
\end{table}

\section{Inductive Node Classification Results}
\label{app:inductive}
For the transductive semi-supervised node classification task, we choose two common used datasets including Reddit and Flikcr \citep{zeng2019graphsaint}. We choose MLP, GCN, APPNP, and C\&S as baselines. The results are shown in Table~\ref{tab:ind1}.

\begin{table}[htb]
\vspace{-0.1in}\centering
\caption{Inductive node classification accuracy (\%).}
\vspace{-0.1in}
\label{tab:ind1}
\resizebox{0.7\textwidth}{!}{%
\begin{tabular}{c|cccccc}
\hline
Method & MLP   & GCN   & APPNP & SGC &  C\&S & \method \\ \hline
Reddit & 62.84 & 93.30 & 94.11 & 93.85 & 95.30 & 95.74   \\
Flickr & 37.87 & 49.20  & 49.40 & 50.58 & 51.46 & 52.29   \\ \hline
\end{tabular}
}
\end{table}

\section{More Efficiency Results}
\label{app:efficiency}
In this section, we show more efficiency results of our proposed \method. We choose the state-of-the-art model APPNP as a base model, and then compare the training time, convergence speed and propagation layers between them.
\subsection{Training Time and Convergence Speed Comparison between \method-Full and APPNP}

\noindent \textbf{Traning Time per epoch.} 
We assume \method-Full has the same number of feature propagation layers with APPNP and find that under different labeling rate the training time is very close. We train each model 5000 epochs on different datasets and report the average training time per epoch that does not include inference and test. As shown in Table \ref{tab:time1}, our training time per epoch is consistently less than APPNP. It is 1.74 times faster on the ogbn-arxiv dataset.  For the small datasets, the input feature dimension is high and feature transformation consumes most of the time, so the improvement is not as large as ogbn-arxiv . Besides, the comparison is based on only one epoch and the assumption that these two models have the same propagation layers. Next, we will show that compared to APPNP, our method has a faster converge speed and needs less feature propagation layers to achieve similar performance.

\begin{table}[htb]
\centering
\caption{Training time (ms) per epoch.}
\label{tab:time1}
\begin{tabular}{c|cccc}
\hline
TIME(ms) & Cora & CiteSeer & Pumbed & ogbn-arxiv \\ \hline
APPNP    & 4.02  & 3.67     & 4.19    & 61.48      \\ 
ALT-OPT  & 3.29  & 2.98     & 3.05    & 35.33       \\ 
MLP	     & 1.97	 & 1.94	   & 1.99	& 13.5 \\ \hline
\end{tabular}
\end{table}

\noindent \textbf{Total training time and Convergence epoch.} 
We measure the convergence speed using the number of epochs at which the model achieves the highest validation accuracy. For our model, the pretraining 100 epochs are also included. This comparison is reasonable because APPNP has the same training parameters with our model and the learning rate is the same. We adopt the same transductive setting as Section~\ref{sec:exp}. The results on Cora and ogbn-arxiv dataset are shown in Table \ref{tab:epoch}.  Our method can converge much faster than APPNP. After 100 epochs fast fully-supervised pretraining, our method only needs tens of iterations to get the best validation accuracy while APPNP needs a few hundreds. Besides, the training epochs of APPNP tend to increase when the label rate increases, while our method tends to decrease. The reason could be that when the label rate increases, the accuracy of label propagation in ALT-OPT also increases, which leads to a faster convergence rate. The total training time is also shown in Table~\ref{tab:epoch}, which shows that ALT-OPT is on average $3.9$ times faster than APPNP.

\begin{table}[htb]
\centering
\caption{The best validation epochs and total training time.}
\label{tab:epoch}
\begin{tabular}{c|cccccc|c}
\hline
Dataset          & \multicolumn{6}{c|}{Cora}               & Ogbn-arxiv \\ \hline
Label rate       & 5    & 10   & 20   & 60   & 30\% & 60\% & 54\%       \\ \hline
APPNP  best epochs          & 283  & 187  & 211  & 283  & 411  & 385  & 342        \\
\method-Full best epochs         & 144  & 151  & 157  & 145  & 121  & 129  & 192        \\
APPNP training time (s)   & 1.14 & 0.75 & 0.85 & 1.14 & 1.65 & 1.55 & 21.0       \\
\method-Full training time (s) & 0.34 & 0.36 & 0.38 & 0.34 & 0.26 & 0.29 & 4.60       \\ \hline
\end{tabular}
\end{table}

\subsection{Number of Feature Propagation Layers.}

Our \method accumulates the pseudo-label $\vF$ during its updating process, it may not require as many propagation layers as APPNP, which would further reduce the training time. To verify this, we conduct experiments to show how the performance changes across different numbers of propagation layers $K$ on Cora dataset with different label rates. The mean accuracy is reported in Table~\ref{tab:layer}. At the label rate 5 and 10, ALT-OPT only needs 1 feature propagation layer to have a comparable result with 10 layer APPNP. For label rate 20, 5 propagation layers can achieve better performance than APPNP. Fewer propagation layers suggest that our \method is more efficient.

\begin{table}[htb]
\centering
\caption{Accuracy(\%) under different feature propagation layer K and different label rates on Cora dataset.}
\label{tab:layer}
\scalebox{0.85}{
\begin{tabular}{c|lllll|ccccc|ccccc}
\hline
Label Rate & \multicolumn{5}{c|}{5}                                                                                                  & \multicolumn{5}{c|}{10}               & \multicolumn{5}{c}{20}                \\ \hline
K          & \multicolumn{1}{c}{1} & \multicolumn{1}{c}{2} & \multicolumn{1}{c}{3} & \multicolumn{1}{c}{5} & \multicolumn{1}{c|}{10} & 1     & 2     & 3     & 5     & 10    & 1     & 2     & 3     & 5     & 10    \\ \hline
APPNP      & 65.49                 & 72.52                 & 74.71                 & 75.64                 & 75.86                   & 71.96 & 77.21 & 78.53 & 79.54 & 80.29 & 77.31 & 80.81 & 81.58 & 82.24 & 82.34 \\
ALT-OPT    & 76.26                 & 76.34                 & 76.42                 & 76.49                 & 76.78                   & 79.59 & 79.59 & 79.69 & 80.01 & 80.66 & 80.76 & 81.66 & 81.95 & 82.48 & 82.66 \\ \hline
\end{tabular}
}
\end{table}

\subsection{Memory Cost of Different Propagation Layers}
Our \method does not need to store hidden states during propagation. The memory cost of \method can be constant with more propagation layers. However, APPNP need to store the hidden states in each propagation layers for backpropagation, and the memory cost grows with more propagation layers. We choose the large ogbn-products dataset for experiment. Both APPNP and \method use 3 transformation layers. We choose different propagation layers, i.e., 10, 20, and 30. The results are shown in Table~\ref{tab:layermem}. The memory cost of \method keep the same with the increase of propagation layers, while the memory of APPNP increase with the increase of propagation layers.

\begin{table}[!htb]
\centering
\caption{Memory Cost (GB) with different propagation layers.}
\label{tab:layermem}
\begin{tabular}{c|ccc}
\hline
Propagation Layer & 10    & 20    & 30    \\ \hline
 APPNP          & 29.15 & 31.31 & 34.43 \\ 
ALT-OPT             & 24.49  & 24.49  & 24.49  \\ \hline
\end{tabular}
\end{table}



